\documentclass[lettersize,journal]{IEEEtran}
\usepackage{amsmath,amsfonts}
\usepackage{algorithmic}
\usepackage{algorithm}
\usepackage{array}
\usepackage[caption=false,font=normalsize,labelfont=sf,textfont=sf]{subfig}
\usepackage{textcomp}
\usepackage{stfloats}
\usepackage{url}
\usepackage{verbatim}

\usepackage{graphicx}
\usepackage{multirow}
\usepackage{booktabs}
\usepackage{bigstrut}
\usepackage{color}
\usepackage{makecell}
\usepackage{hyperref}

\hypersetup{
	colorlinks=true,
	linkcolor=black,
	citecolor=green,
}

\begin{document}

\title{Collaborative Camouflaged Object Detection:\\ A Large-Scale Dataset and Benchmark\\}

\author{Cong Zhang, Hongbo~Bi,~\IEEEmembership{Member,~IEEE,}~Tian-Zhu Xiang,~Ranwan Wu,~Jinghui Tong,~Xiufang Wang
\thanks{Manuscript was accepted by IEEE Transactions on Neural Networks and Learning Systems (TNNLS).}
\thanks{
Cong Zhang, Hongbo Bi, Ranwan Wu, Jinghui Tong and Xiufang Wang are with the Department of Electrical Information Engineering, Northeast Petroleum University, Daqing, 163318, China (congzhang98@163.com, bhbdq@126.com, wuranwan2020@sina.com, tongjinghui0601@163.com, wxfdqpi@163.com). 
 
Tian-Zhu Xiang is with the Inception Institute of Artificial Intelligence and G42 Bayanat, Abu Dhabi, UAE (tianzhu.xiang19@gmail.com).

Co-corresponding authors: Hongbo Bi and Tian-Zhu Xiang.
}
}

\markboth{Journal of \LaTeX\ Class Files,~Vol.~14, No.~8, August~2021}%
{Shell \MakeLowercase{\textit{et al.}}: A Sample Article Using IEEEtran.cls for IEEE Journals}


\maketitle

\begin{abstract}
In this paper, we provide a comprehensive study on a new task called collaborative camouflaged object detection (CoCOD), which aims to simultaneously detect camouflaged objects with the same properties from a group of relevant images. To this end, we meticulously construct the first large-scale dataset, termed CoCOD8K, which consists of  8,528 high-quality and elaborately selected images with object mask annotations, covering 5 superclasses and 70 subclasses. The dataset spans a wide range of natural and artificial camouflage scenes with diverse object appearances and backgrounds, making it a very challenging dataset for CoCOD. Besides, we propose the first baseline model for CoCOD, named bilateral-branch network (BBNet), which explores and aggregates co-camouflaged cues within a single image and between images within a group, respectively, for accurate camouflaged object detection in given images. This is implemented by an inter-image collaborative feature exploration (CFE) module, an intra-image object feature search (OFS) module, and a local-global refinement (LGR) module. We benchmark 18 state-of-the-art models, including 12 COD algorithms and 6 CoSOD algorithms, on the proposed CoCOD8K dataset under 5 widely used evaluation metrics. Extensive experiments demonstrate the effectiveness of the proposed method and the significantly superior performance compared to other competitors. We hope that our proposed dataset and model will boost growth in the COD community. The dataset, model, and results will be available at: \textcolor{blue}{ \url{https://github.com/zc199823/BBNet--CoCOD}}.
\end{abstract}


\begin{figure}[htb]
	\centering
	\includegraphics[width=1\linewidth]{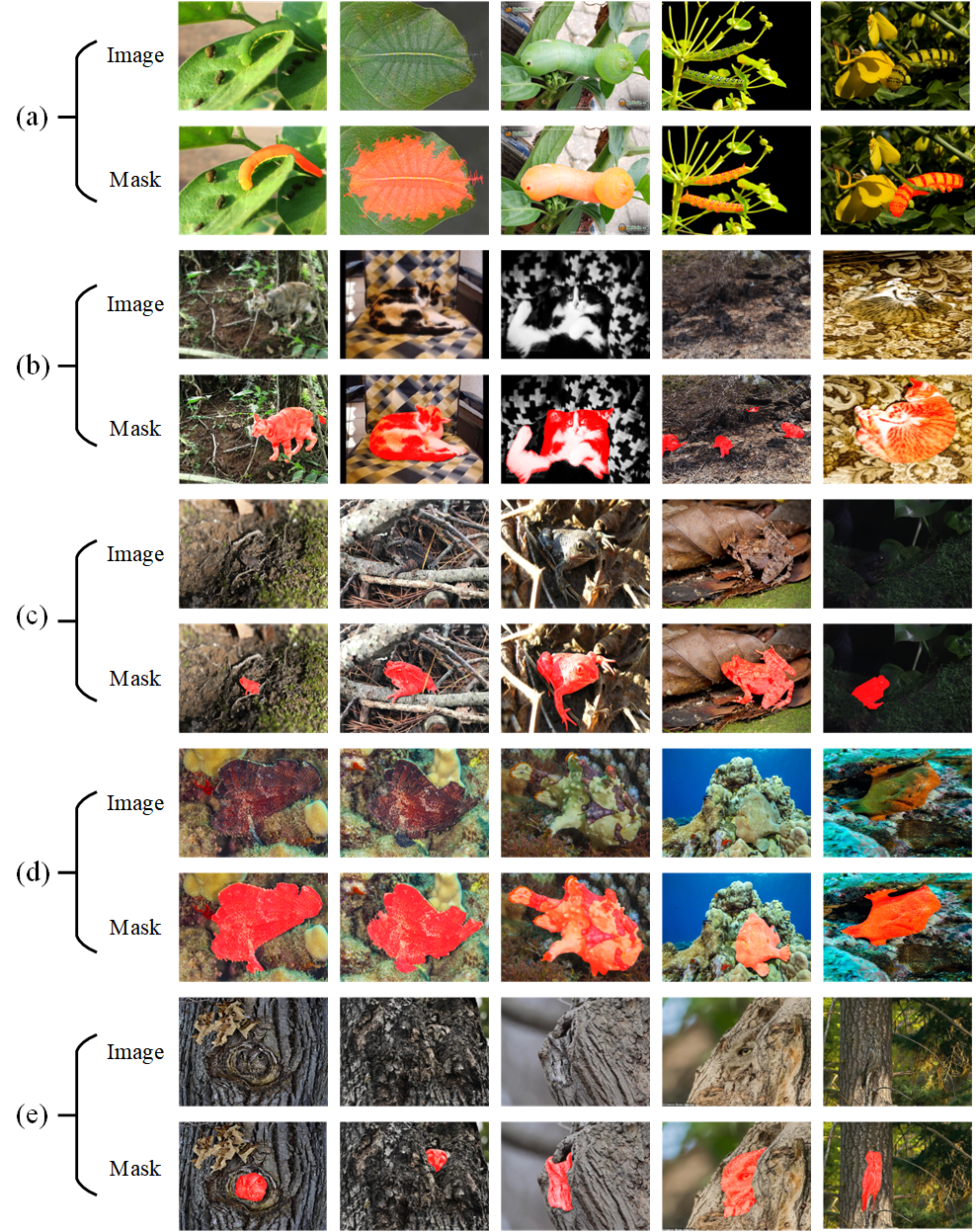}
	\caption{Illustration of collaborative camouflaged object detection. Five image groups are from the proposed CoCOD8K datasets, which are (a) crawling insects, (b) cats, (c) toads, (d) frogfish, and (e) owls, respectively.}
	\label{fig:cocod_dataset}
\end{figure}

\section{Introduction}
\IEEEPARstart{I}{n} recent years, camouflaged object detection (COD), segmenting objects that are visually blended in with their background, has attracted increasing attention from the computer vision community~\cite{fan2023advances, dong2023unified}. 
COD has shown great potential for a wide range of downstream applications, such as wildlife conservation~\cite{nafus2015hiding}, medical image segmentation~\cite{li2022TVS, li2021mvdi25k}, agricultural pest detection~\cite{chahl2018bioinspired}, and artistic creation~\cite{li2022location}.
Compared with generic object detection, COD is a very challenging task due to the high intrinsic similarities between the camouflaged objects and their background.

Thanks to the construction of large-scale camouflaged object detection datasets such as CAMO~\cite{le2019anabranch}, COD10K~\cite{fan2020camouflaged}, and NC4K~\cite{lv2021simultaneously}, COD has received extensive studies. An increasing number of recent works have been proposed to leverage deep learning techniques for camouflaged object detection, such as coarse-to-fine learning method (\textit{e.g.}, SINet~\cite{fan2020camouflaged} and C2FNet~\cite{sun2021context}), joint-learning method (\textit{e.g.}, LSR~\cite{lv2021simultaneously} and BGNet~\cite{sun2022boundary}), uncertainty-aware learning method (\textit{e.g.}, UGTR~\cite{yang2021uncertainty} and ZoomNet~\cite{pang2022zoom}), multi-modal information fusion (\textit{e.g.}, frequency~\cite{zhong2022detecting} and temporal cues~\cite{cheng2022implicit}), and generative method (\textit{e.g.}, diffCOD~\cite{chen2023diffusion}). 

Although these sophisticated approaches have achieved great progress, they still struggle to extract discriminative camouflaged object features from highly similar and easily confused backgrounds. 
We argue that, in many cases, it is difficult to dig out enough valuable clues from only single images to discriminate camouflaged objects. 
As we know, when the human observes difficult-to-detect targets, the visual system will instinctively sweep across such a collection of related images to search for more valuable clues~\cite{zhang2016detection}. 
Inspired by this visual mechanism, we introduce a novel camouflaged object detection task, namely collaborative camouflaged object detection (CoCOD), which jointly segments the same camouflaged object (or camouflaged objects of the same class) in multiple distinct images. 
Actually, studies~\cite{li2018deep} have shown that jointly segmenting multiple images can achieve better accuracy than segmenting them independently in the general object segmentation field.  
The commonality shared across the images and complement cues from multiple images of different scenarios may significantly facilitate the success of accurate camouflaged object detection. 

As there exist no corresponding datasets for the proposed new task, in this paper, we meticulously construct a collaborative camouflaged object dataset based on the existing COD dataset, shown in Fig.~\ref{fig:cocod_dataset}, which extends the COD from the individual image to the image group, facilitating novel and effective solutions for COD. However, it is noted that the same camouflaged objects in multiple images will have different degrees of variability in terms of scale, appearance, pose, viewpoint, background, and object (self-)occlusion, which brings challenges to CoCOD. How to explore and aggregate effective subtle cues from different images for accurate camouflaged object detection deserves more effort.

\begin{figure}[tb]
	\centering
	\includegraphics[width=1\linewidth]{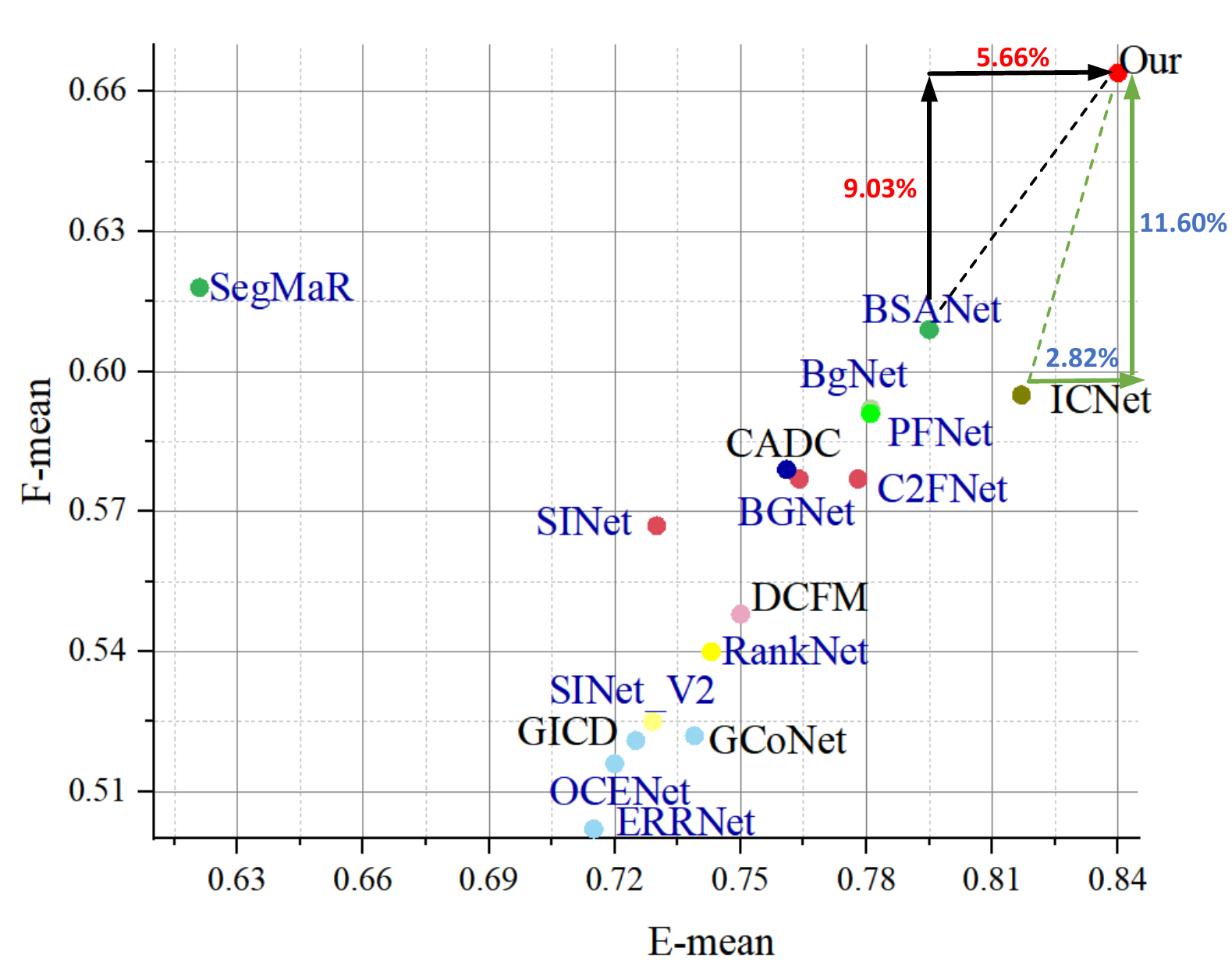}
	\caption{Mean F-measure ($F_{mean}$) vs. Mean E-measure ($E_{mean}$) of top 16 methods (11 COD models shown in blue and 5 CoSOD models shown in black) from 18 SOTA methods and our method on the proposed CoCOD8K dataset. Note that JCOSOD~\cite{li2021uncertainty} and CoEGNet~\cite{fan2021re} are not shown in this figure since their $F_{mean}$ and $E_{mean}$ are lower than 0.5 and 0.6 respectively. $F_{mean}$ is a comprehensive metric to evaluate the weighted harmonic mean between precision and recall of predicted results. $E_{mean}$ aims to evaluate the both local and global similarity between the predictions and ground truth simultaneously. We can see that our method achieves a remarkable performance improvement compared to other competitors.}
\label{fig:vs_f_e}
\end{figure}

To tackle these challenges, based on the proposed dataset, we propose a simple but effective baseline model, termed bilateral-branch network (BBN) for CoCOD, which explores subtle camouflaged cues within a single image and between images within a group, respectively, and then aggregates them for accurate camouflaged object detection in given images. Specifically, we design a collaborative feature exploration (CFE) module based on feature shuffle and multi-view strategies to excavate consensus semantics of co-camouflaged objects within image groups for locating the co-camouflaged objects. 
In addition, we also devise an intra-image object feature search (OFS) module to capture more fine cues of camouflaged objects within the image by focusing on a single image to deeply observe the relationship between the camouflaged object and the background. 
Then, the inter-image and intra-image camouflage object features are integrated, and a local-global refinement (LGR) module is proposed to improve feature representation by exploring local and global cues for co-camouflaged object prediction.
As shown in Fig.~\ref{fig:vs_f_e}, we can see the proposed method achieves a significant performance improvement compared to other existing state-of-the-art COD models and CoSOD models. The contributions of this work can be summarized as:

\begin{enumerate}[]
	\item \textbf{CoCOD8K Dataset:} We propose the first large-scale and challenging CoCOD dataset, termed CoCOD8K, which contains 8,528 high-resolution camouflaged images paired with corresponding binary mask annotations. This work provides a foundation for the CoCOD field and is expected to give a strong boost to growth in the COD community.

    \item  \textbf{CoCOD Model:} We propose a novel strong baseline, dubbed bilateral-branch network (BBN), for CoCOD, which excavates and aggregates camouflaged object features from the inter-image collaborative feature exploration branch and intra-image object feature search branch, respectively, and strengthens co-camouflaged feature representation by local-global feature refinement for co-camouflaged object detection. 

    \item \textbf{Empirical Contribution:} We offer a rigorous evaluation of 18 state-of-the-art models (12 COD algorithms and 6 CoSOD algorithms) under 5 widely used evaluation metrics, which demonstrates the effectiveness of the proposed method. Comprehensive benchmarking results show that the proposed BBNet significantly outperforms other competitors, making it an effective solution for the CoCOD task.
 
\end{enumerate}

\section{Related work} 
\label{sec:Relatedwork}

\textbf{Camouflaged Object Detection}. 
Camouflaged object detection aims to discover camouflaged objects from their high-similarity surroundings. 
Early methods often adopt hand-crafted features to detect camouflaged objects, such as texture~\cite{li2017foreground}, 3D convexity~\cite{pan2011study}, motion information~\cite{hall2013camouflage,li2018fusion} and Gaussian
mixture model~\cite{gallego2014foreground}, showing the limited capability of spotting camouflaged objects in complex backgrounds.

To this end, recently, numerous deep learning-based models have been proposed to detect camouflaged objects and have achieved compelling performance~\cite{bi2021rethinking}. Through thorough analysis, these methods can be roughly divided into five strategies: coarse-to-fine learning strategy, joint-learning strategy, uncertainty-aware learning strategy, multi-modal information fusion strategy, and generative strategy. 
a) \textit{Coarse-to-fine learning strategy} often integrates global prediction (for coarse location) and local refinement (for accurate segmentation). 
Most of the methods are proposed to fuse multi-scale features effectively, such as partial decoder component~\cite{fan2020camouflaged,wang2021d}, neighbor connection decoder and group-reversal attention~\cite{fan2022concealed}, X-shaped connection~\cite{zhuge2022cubenet}, attention-induced cross-level fusion~\cite{sun2021context}, distraction mining~\cite{mei2021camouflaged}, texture-aware refinement~\cite{ren2021deep}, object magnification and iterative refinement~\cite{jia2022segment}, and feature shrinkage pyramid~\cite{huang2023feature}.
b) \textit{Joint-learning strategy} introduces other tasks as auxiliary knowledge and facilitates the camouflaged feature exploration by multi-task collaborative learning, such as classification~\cite{le2019anabranch}, localization and ranking~\cite{lv2021simultaneously}, texture perception~\cite{zhu2021inferring}, gradient supervision~\cite{ji2022deep}, boundary detection~\cite{sun2022boundary,zhu2022can,zhai2021mutual,ji2022fast}, and saliency learning~\cite{li2021uncertainty}.
c) \textit{Uncertainty-aware learning strategy} introduces the confidence-aware learning to model the confidence of model predictions, which facilitates the model to learn from difficult samples and improves its robustness, such as uncertainty-guided transformer reasoning~\cite{yang2021uncertainty}, aleatoric uncertainty modeling~\cite{liu2022modeling}, uncertainty-based regularization constraint~\cite{pang2022zoom}. 
d) \textit{Multi-modal information fusion strategy} resorts to multi-source/multi-modal information to improve the detection of camouflaged objects, such as frequency information~\cite{zhong2022detecting}, temporal information~\cite{cheng2022implicit} and depth cues~\cite{wu2022source}. 
(e) \textit{Generative strategy} casts camouflaged object segmentation as an image-conditioned mask generation problem, via generative learning frameworks, such as the diffusion model~\cite{chen2023diffusion, chen2023camodiffusion}. 

Nevertheless, to our knowledge, existing COD methods only explore camouflaged object features within a single image, which is information-limited and easy to get into trouble for complex and confusing scenarios. To address this issue, we propose collaborative camouflaged object detection, which simultaneously segments the same camouflaged object from multiple relevant images. By observing multiple relevant images, models can explore and aggregate more valuable cues for accurate camouflaged object detection, which will open up a new opportunity for COD. 

\textbf{COD Datasets}.
The dataset is the foundation of deep learning-based camouflaged object detection. Tab.~\ref{tab:alldataset} lists five available image datasets specifically for camouflaged object detection, detailed as follows.
\begin{itemize}
    \item CHAMELEON~\cite{SkurowskiCHAM} is the first natural camouflage dataset, which only consists of 76 images collected from Google search by the keyword ``camouflaged animal". 
    \item CPD1K~\cite{liu2021integrating} is the first artificial camouflage dataset dictated for the camouflaged people segmentation task, which contains 1,000 images of camouflaged soldiers obtained from videos.
    \item CAMO~\cite{le2019anabranch} contains 2,500 images covering eight categories (both natural and artificial camouflaged objects), where 2,000 images are for training, and the remaining 500 images are for testing.
    \item COD10K~\cite{fan2020camouflaged} is currently the largest training and testing benchmark dataset for camouflaged object detection, which contains 10,000 images covering 10 super-categories and 78 sub-categories. 
    \item NC4K~\cite{lv2021simultaneously} is the largest testing dataset with 4,121 images for COD, which is commonly used to evaluate the generalization ability of models.
\end{itemize}

The above datasets are all based on single images.
Therefore, to facilitate research on CoCOD, we classify, count, and filter existing COD datasets to construct a large-scale dataset specially for the CoCOD task, namely CoCOD8K, which contains 8,528 images covering 5 super-categories (that is, insect, land, sea, fly, and amphibious) and 70 sub-categories.

\begin{table}[tb]
	\centering
	\caption{Five commonly-used COD benchmark datasets. We list some dataset information such as year, publication (Pub.), dataset number (Num. for training and testing), number of object types in the datasets (Class), and the camouflage type in the dataset (Type).}
	\renewcommand{\arraystretch}{1.65}	\renewcommand{\tabcolsep}{0.6mm}
	\begin{tabular}{c|c|c|c|c|c|c|c}
		\toprule
		\multirow{2}[4]{*}{Dataset} & \multirow{2}[4]{*}{~~Year~~} & \multirow{2}[4]{*}{~~Pub.~~} & \multirow{2}[4]{*}{~Class~} & \multicolumn{2}{c|}{Num.} &  \multicolumn{2}{c}{Type} \bigstrut\\
		\cline{5-6}\cline{7-8}          &       &     &      & ~Train~ & ~Test~    & \multicolumn{1}{p{4em}|}{Artificial} & \multicolumn{1}{p{4em}}{Natural} \bigstrut\\ 
		\hline
		{\scriptsize CHAMELEON} & 2018  & Unpub. & \# &  0     & 76     &       &  $\surd$ \bigstrut[t]\\
		CPD1K & 2018  & SPL  & 2  & 0     & 1,000   &  $\surd$     &  \\
		CAMO  & 2019  & CVIU  & 8  & 2,000  & 500    &  $\surd$     &  $\surd$ \\
		COD10K & 2021  & TPAMI  & 78   & 6,000  & 4,000  &  $\surd$     & $\surd$ \\
		NC4K  & 2021  & CVPR   & \#   & 0     & 4,121   &  $\surd$     &  $\surd$ \bigstrut[b] \\ \hline
		CoCOD8K  & 2023 & - & 70 & 5,933 & 2,595 & $\surd$ & $\surd$ \\
		\bottomrule
	\end{tabular}%
	\label{tab:alldataset}%
\end{table}%

\textbf{Co-salient Object Detection}. 
Co-salient object detection (CoSOD) is to simultaneously detect common salient objects from a group of relevant images~\cite{zheng2023memory}. Early methods attempt to explore the semantic attributes shared among images in a group to model the inter-image correspondence for co-salient object detection in different ways, such as regional histograms and contrasts~\cite{liu2013co}, self-adaptive weights~\cite{cao2014self}, metric learning~\cite{han2017unified}, efficient manifold ranking~\cite{li2014efficient} and base reconstruction~\cite{cao2014co}, based on handcrafted heuristic characteristics~\cite{chang2011co,fu2013cluster} (\textit{e.g.}, color, texture, and contour).

Recently, deep learning-based CoSOD methods have demonstrated compelling performance by learning semantic representations of co-salient objects jointly in a supervised manner~\cite{ge2022tcnet, ge2023gsnnet}. 
For instance, Zhang \textit{et al.} \cite{zhang2020gradient} introduced a gradient-induced model (GICD) to mine more discriminative co-salient features using the feedback gradient information.
Jin \textit{et al.} \cite{jin2020icnet} proposed an intra-saliency correlation network (ICNet) to explore the relationship among saliency maps predicted by any SOD frameworks. 
Fan \textit{et al.}~\cite{fan2021re} extended the prior edge-guided saliency detection network by introducing a co-attention projection strategy for co-salient object detection.
Zhang \textit{et al.} \cite{zhang2021summarize} presented a consensus-aware dynamic convolution network (CADC) that exploits the self-attention mechanism to explore cross-image consensus representation. 
More recently, Fan \textit{et al.}~\cite{fan2021group} designed a group collaborative learning network (GCoNet) to simultaneously mine intra-group feature and inter-group collaborative salient cues between different image groups. 
Yu \textit{et al.}~\cite{yu2022democracy} introduced a democratic prototype generation model (DCFM) which excavates sufficient co-salient regions by generating democratic response maps, for complete co-saliency prediction.

It is worth noting that CoCOD is significantly different from CoSOD. Although both tasks aim to segment objects from multiple images simultaneously, CoSOD is to discover those salient and easily observed objects~\cite{fan2022salient, huang2022scribble}, whereas CoCOD is to deal with objects with a lower probability of being noticed (\textit{i.e.}, inconspicuous objects), which is far more challenging. 
Unlike the above methods, we propose a bilateral branch network to explore and integrate subtle consensus cues of similar camouflaged objects from two perspectives: inter-image and intra-image, respectively, for collaborative camouflaged object detection, showing significant performance improvement, compared to directly segmenting camouflaged objects in each image separately.

\section{CoCOD8K Dataset}\label{sec:ourdataset}

To facilitate research on collaborative camouflaged object detection, we construct the first large-scale and challenging CoCOD8K dataset, which contains 8,528 images with diverse scenes, covering 5 super-classes and 70 sub-classes, detailed in this section. 

\begin{figure}[tb]
	\centering
	\includegraphics[width=0.95\linewidth]{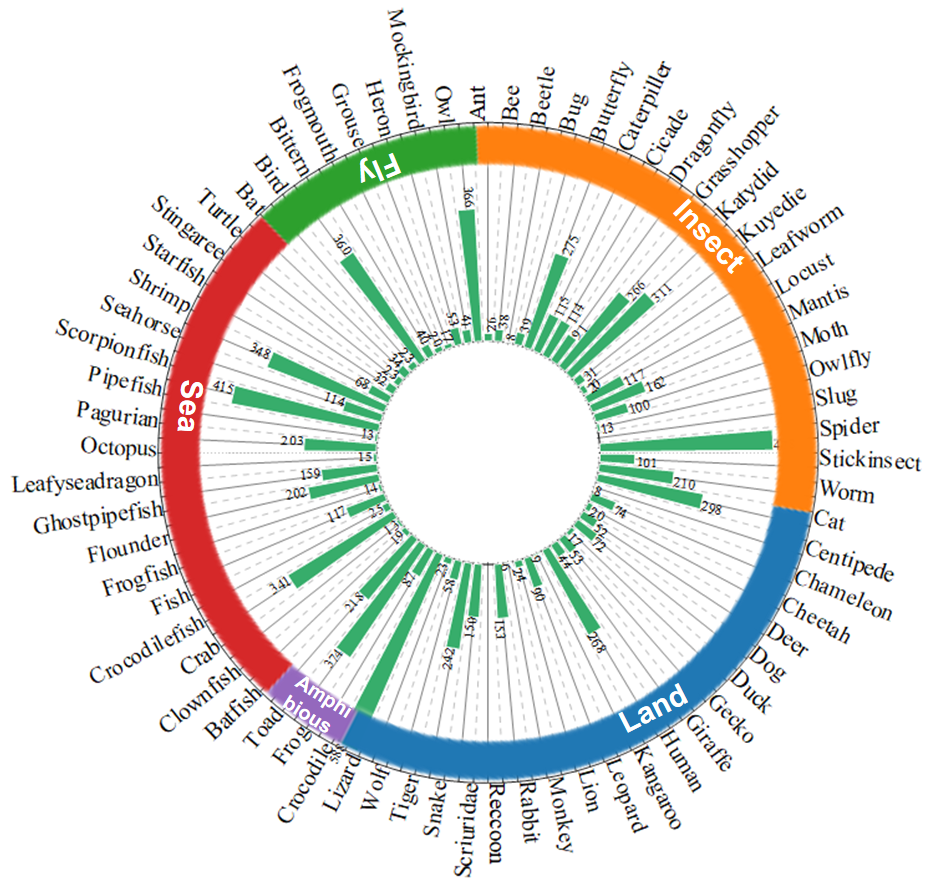}
	\caption{Taxonomic structure of our proposed dataset, which illustrates the histogram distribution for the 5 super-classes and 70 sub-classes in our CoCOD8K.}
	\label{fig:toxonomic}
\end{figure}

\subsection{Image Collection and Annotation}
Different from COD, CoCOD aims to simultaneously detect co-camouflaged objects from a group of relevant images, which remains an under-explored topic. Besides, to our knowledge, there is no large-scale annotation dataset specifically for CoCOD, which seriously hinders the development of CoCOD. To this end, we build the first large-scale CoCOD8K dataset by re-labeling and re-organizing four existing COD datasets, \textit{i.e.}, CHAMELEON, CAMO, COD10K, and NC4K. 

\begin{figure}[ht]
\centering
\includegraphics[width=0.98\linewidth]{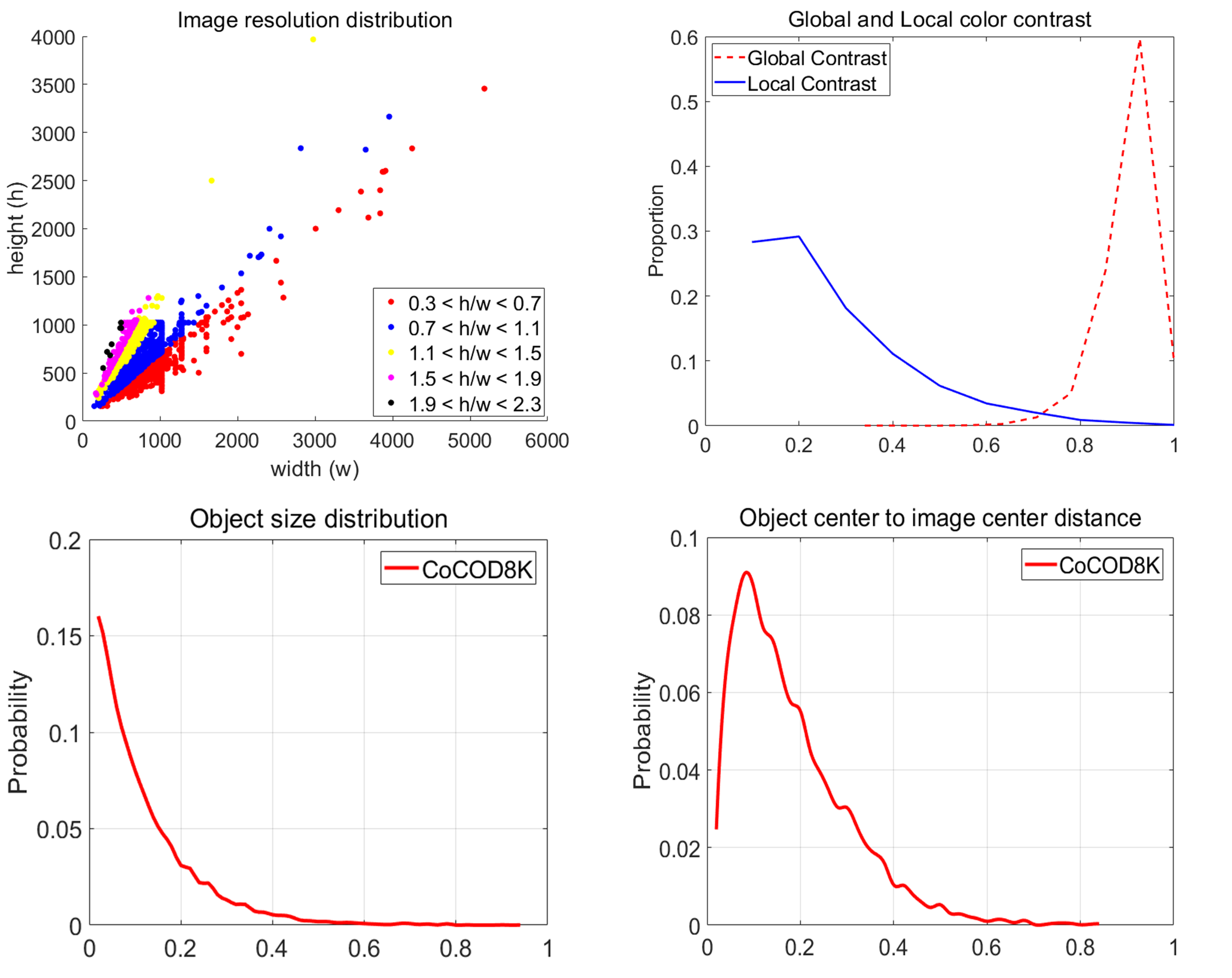}
\caption{Dataset features and statistics for CoCOD8K. (a) Image resolution distribution. (b) Global/local contrast distribution. (c) Object size distribution. (d) Object center to image center distance.}
\label{fig:ISD}
\end{figure}

\subsubsection{Image annotation} As some of the existing COD datasets do not provide corresponding category annotation, we label the object categories in each image by biological attributes. 
Specifically, we first classify these images into 5 super-classes, \textit{i.e.}, Land, Sea, Fly, Amphibians, and Insects, according to the living environment of the camouflaged objects. Generally, for the task of collaborative object detection, an image group contains only one class of objects. Therefore, we further categorize these data into 70 sub-classes (such as people, owls, crabs, toads, and worms) based on the biological properties of the camouflaged objects. Fig.~\ref{fig:toxonomic} shows the taxonomic structure of our proposed dataset. 
Finally, for each image in our dataset, we provide 2 types of annotations, including categories (super-class and sub-class categories) and object masks.

\begin{figure*}[tb]
	\centering
	\includegraphics[width=1\linewidth]{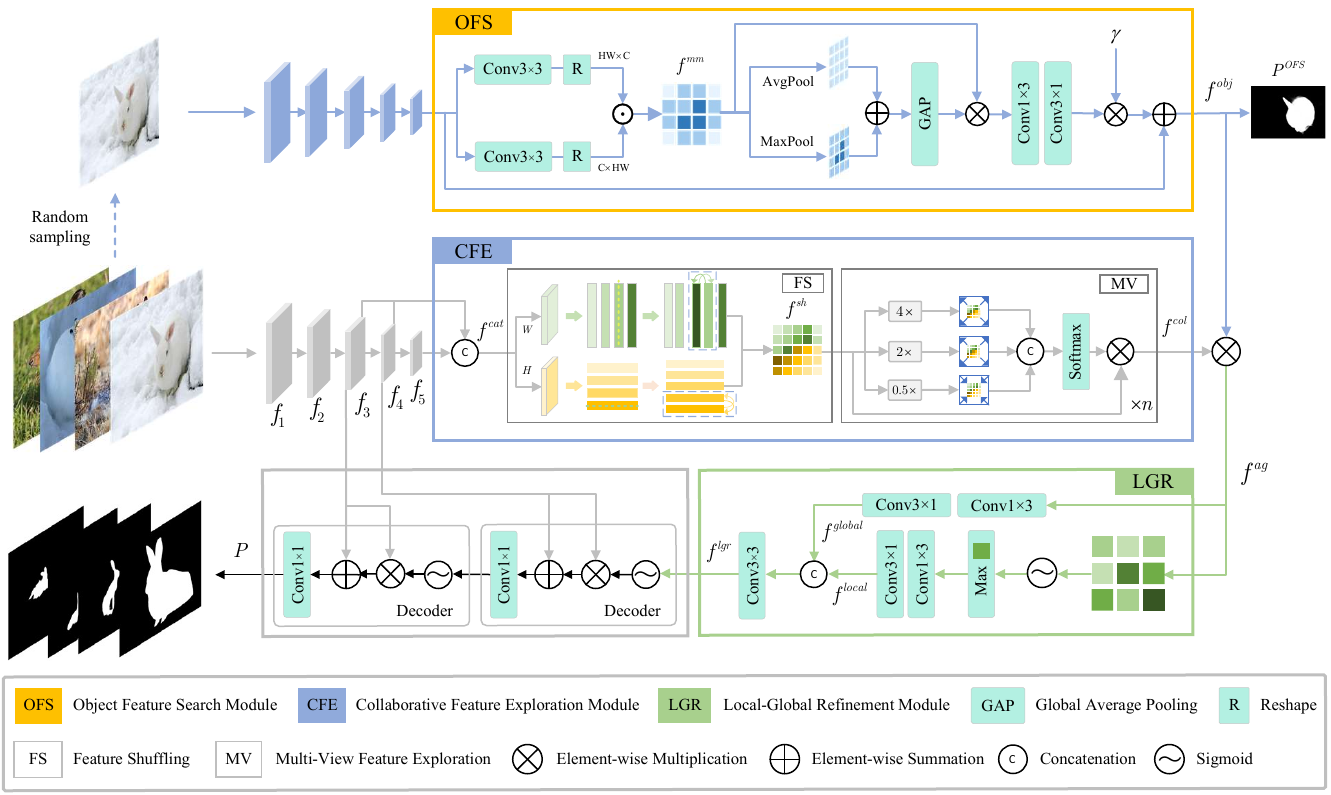} 
	\caption{An overview of our proposed BBNet. Specifically, our model contains three main modules, \textit{i.e.}, collaborative feature extraction (CFE), object feature search (OFS), and local-global refinement (LGR). The CFE and OFS modules are designed to explore subtle camouflaged object cues ($f^{col}$ and $f^{obj}$) from within image groups and within images, respectively. These two features are then integrated, and the LGR module is devised to enhance the representation of co-camouflaged features from both global and local perspectives. 
	Combining the co-camouflaged feature ($f^{lgr}$) with backbone features ($f_3$ and $f_4$) via two simple decoders, the model provides accurate camouflaged object predictions from the given images. $\times n$ denotes $n$ iterations.
	}
	\label{fig:architecture}
\end{figure*}

\subsubsection{Image filtering} Through thoroughly analyzing the existing COD dataset, we find that some images contain two or more camouflaged objects of different classes, which are unsuitable for the CoCOD task. Therefore, we carefully examine the entire dataset and filter out these controversial images. Furthermore, to ensure adequate training data for deep models, we filter out the image groups with less than 5 images. Finally, we obtained a total of 8,528 images for CoCOD and split them into 5,933 images for training and 2,595 images for testing. The details are listed in Tab.~\ref{tab:dataset}.

\begin{table}[tb]
  \centering
  \caption{The details of each sub-dataset in CoCOD8K dataset}
  \renewcommand{\arraystretch}{1.35}	\renewcommand{\tabcolsep}{4mm}
    \begin{tabular}{c|c|ccc}
    \toprule
    \multirow{2}[4]{*}{Sub-dataset} & \multirow{2}[4]{*}{\makecell[c]{Num. of \\Classes}} & \multicolumn{3}{c}{Num.} \bigstrut\\
\cline{3-5}          &     & Training & Testing  & Total \bigstrut\\
    \midrule
    Land  & 21    & 1632  & 622   & 2254 \bigstrut[t]\\
    Insect & 20    & 1654  & 866   & 2520 \\
    Sea   & 18    & 1521  & 634   & 2155 \\
    Fly   & 8     & 638   & 282   & 920 \\
    Amphibious & 3     & 488   & 191   & 679 \bigstrut[b]\\
    \bottomrule
    \end{tabular}%
  \label{tab:dataset}%
\end{table}%

\subsection{Dataset Features and Statistics} 
Fig.~\ref{fig:ISD} summarizes some statistical features for the CoCOD8K dataset.

\begin{itemize}
    \item Resolution distribution. Fig.~\ref{fig:ISD} (a) shows the image resolution distribution of the constructed CoCOD8K, which includes a large number of high-resolution images. In general, high-resolution images typically contain more object detail, which benefits model training and prediction performance~\cite{zhang2021looking}. 
    \item Global/local contrast. Inspired by~\cite{li2014secrets}, we adopt global and local color contrast distribution to evaluate the difficulty of object detection in our dataset. As shown in Fig.~\ref{fig:ISD} (b), most camouflaged objects are with low local contrast, that is, the objects are well hidden in the surroundings and show high similarities with the local background. Thus, objects in CoCOD8K are very challenging for accurate detection and segmentation. 
    \item Object size. Fig.~\ref{fig:ISD} (c) shows the normalized object size distribution of our CoCOD8K dataset, that is, the proportion of pixels of camouflaged objects to input images. As can be seen, the object size distribution varies over a broad range, from 2\% to 94\%. In addition, in our dataset, small and medium-sized objects account for a higher proportion. 
    \item Center bias. Fig.~\ref{fig:ISD} (d) shows the distance of the object center to the image center. As can be seen, the center bias distribution is from 0.02 to 0.84 (avg.: 0.43), indicating that our collected CoCOD8K dataset suffers from less center bias.
\end{itemize}

\section{Our approach}\label{sec:approach}
In this section, we will introduce our proposed BBNet, which explores and integrates intra-image and inter-image camouflaged consensus cues for CoCOD.

\subsection{Overall Architecture}
Fig.~\ref{fig:architecture} illustrates the overall architecture of the proposed BBNet. For the inter-image branch, the backbone features are extracted from the input image group and then fed into the designed collaborative feature exploration (CFE) module to mine consensus semantics of co-camouflaged objects based on feature shuffle and multi-view strategies. Besides, we design an intra-image branch that adopts the devised object feature search (OFS) module to extract fine-grained camouflaged object cues from the backbone features within a single image that is randomly sampled from the image group. After that, the two branch features are integrated, and then a local-global refinement (LGR) module is proposed to further strengthen the feature representation of co-camouflaged objects by local and global feature exploration. Finally, a simple decoder is adopted to integrate the enhanced co-camouflaged feature with the backbone features for co-camouflaged object prediction. 

\subsection{Collaborative Feature Exploration (CFE)} 
Camouflaged objects tend to be highly similar to their backgrounds, and these objects in the image group often show large differences in object appearance (\textit{i.e.}, object size, shape, texture, and scale), so it is often difficult to distinguish the camouflaged object features which are inconspicuous through simple observation. In this work, we propose a collaborative feature exploration (CFE) module, which meticulously mines the clues of camouflaged objects by feature shuffling and deep observation with a multi-view strategy. 

To locate co-camouflaged objects from a group of images, we take the third-layer backbone features as the bifurcation point and combine three high-level features to excavate consensus semantic information. Specifically, we first utilize channel concatenation operations to integrate 3$^{rd}$, 4$^{th}$, and 5$^{th}$ level features $(f_{3},f_{4},f_{5})$ to obtain the concatenate feature $f^{cat} \in \mathbb{R}^{B\times C\times H \times W}$, where $B$, $C$, $H$ and $W$ denote the batch size, channel, height, and width of the features, respectively. 
Then, we design a feature shuffling operation based on the W and H dimensions to break the pixel relationship between the original object and the background, thereby obtaining more valuable information and highlighting co-camouflaged objects. 
More specifically, we divide $f^{cat}$ into two feature sets $f^{cat}_h \in \mathbb{R}^{B\times C\times 2\times H/2 \times W}$ and $f^{cat}_w \in \mathbb{R}^{B\times C\times H\times 2\times W/2}$ according to the 3$^{rd}$ and 4$^{th}$ dimensions, respectively. After that, we shuffle and exchange the split dimensions, respectively, and obtain the exchanged features $f^{sh}_h \in \mathbb{R}^{B\times C\times H/2\times 2 \times W}$ and $f^{sh}_w \in \mathbb{R}^{B\times C\times H\times W/2\times 2}$. Next, we combine these two features using a dimension concatenation operation along the H and W dimensions to obtain the final shuffled feature $f^{sh}$.

\begin{figure}[tb]
\centering
\includegraphics[width=1\linewidth]{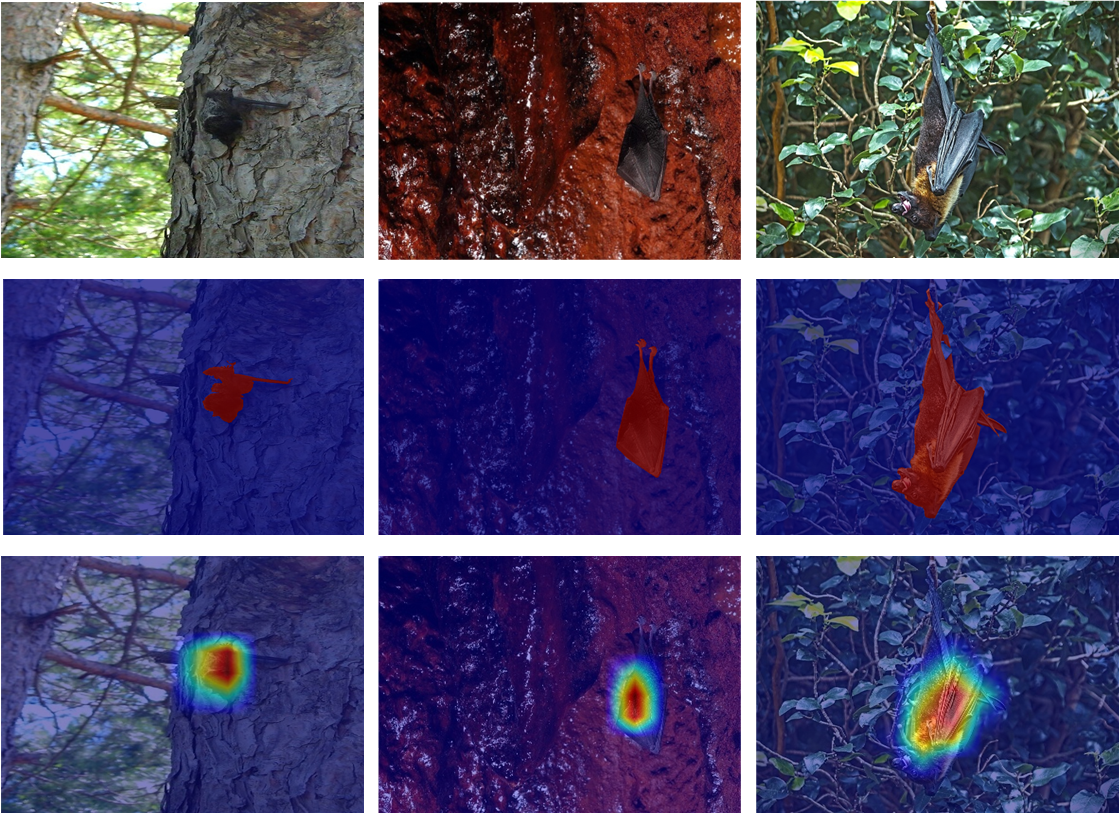} 
\caption{The effectiveness of the model: backbone with CFE module. From top to bottom are input images, corresponding ground truths, and prediction maps, respectively.}
\label{fig:cfe}
\end{figure}

Note that dimensional destruction destroys the potential pixel layout of the original images, and may lead to incomplete collaborative feature extraction while providing new features. To this end, we present a multi-view feature exploration approach to mimic the human visual observation mechanism for a complete collaborative feature exploration, that is, observing co-camouflaged features based on zoom-in and zoom-out receptive fields, respectively. Specifically, we use a parallel structure to scale the dimensional-shuffled feature maps by 0.5 times, 2 times, and 4 times, respectively, after dimensional shuffling, and then concatenate them to obtain the camouflaged consensus semantic of the input image group, which can be formulated as: 
\begin{equation}
\left \{
\begin{aligned}
f_\downarrow^{\alpha} & = \delta^{\alpha}_\downarrow (f^{sh}), \alpha = 0.5, \\
f_\uparrow^{\beta} & = \delta^{\beta}_\uparrow (f^{sh}), \beta \in \{2, 4\}, \\ 
f^{mv} & = f^{sh} \otimes {\rm S} ( {\rm Cat} (f_\downarrow^{\alpha}, f_\uparrow^{\beta}) ),
\end{aligned}
\right.
\end{equation}
where $\rm Cat(\cdot)$ denotes the concatenation operation, and ${\rm S} (\cdot)$ denotes the Softmax function. $\delta^{\alpha}_\downarrow$ and $\delta^{\beta}_\uparrow$ represents a downsampling (\textit{i.e.}, 0.5 times) operation and a upsampling (\textit{e.g.}, 2 and 4 times) operation, respectively. $\otimes$ is element-wise multiplication.  
In addition, to strengthen the co-camouflaged features, we perform the multi-view operation on $f^{mv}$ iteratively and generate the final co-camouflaged feature $f^{col}$. Fig.~\ref{fig:cfe} shows some predicted co-camouflage maps output from the proposed CFE module, demonstrating the ability to accurately localize the co-camouflaged objects from the input image group. 

\subsection{Object Feature Search (OFS)}
Compared with other collaborative detection tasks (\textit{e.g.}, co-salient object detection), objects in CoCOD are highly similar to their backgrounds, which brings challenges to object segmentation. We argue that relying solely on inter-image feature exploration (\textit{i.e.}, the CFE branch) may not be able to completely separate objects and backgrounds. Therefore, we introduce another branch in which we propose an object feature search (OFS) module to capture intra-image camouflage cues. Specifically, we randomly sample an image from the image group as input to extract backbone features ($f^{sin}$), and then perform dimension compression operations on the input backbone features to obtain two sets of features, \textit{i.e.}, $f^{sin}_1 \in \mathbb{R}^{B\times HW\times C}$ and $f^{sin}_2 \in \mathbb{R}^{B\times C\times HW}$, respectively. Next, we use matrix multiplication to strengthen intra-image features, which is denoted as: 
\begin{equation}
\left\{
    \begin{aligned}
		f^{sin}_1 & = {\rm R}( {\rm F}_{Conv3\times3} (f^{sin}) ), \\ 
		f^{sin}_2 & = {\rm R}( {\rm F}_{Conv3\times3} (f^{sin}) ), \\ 
		f^{mm} & = f^{sin}_1 \odot f^{sin}_2,
	\end{aligned}
\right.
\end{equation}
where ${\rm F}_{Conv3\times3} (\cdot)$ denotes a 3$\times$3 convolutional layer, and $R (\cdot)$ denotes dimension transformation. $f^{mm}$ is the output feature, and $\odot$ is a matrix multiplication operation. 

\begin{figure}[tb]
\centering
\includegraphics[width=1\linewidth]{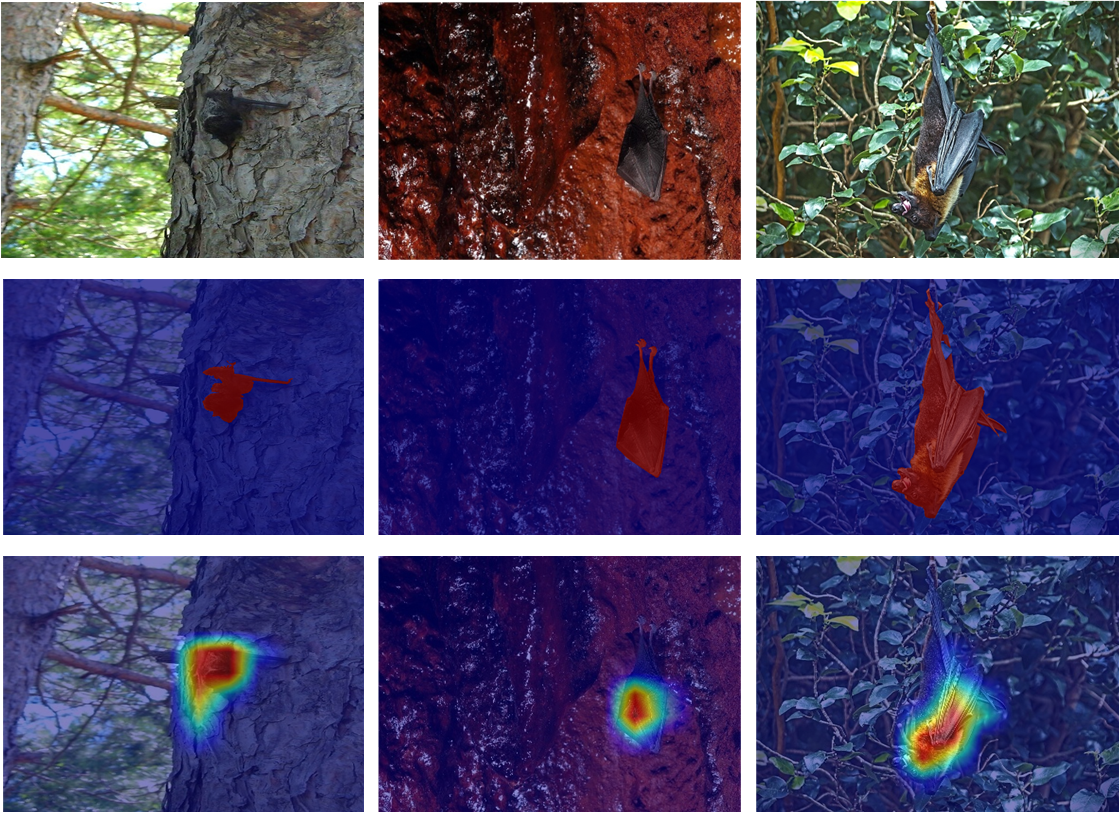} 
\caption{The effectiveness of the model: backbone with CFE and OFS modules. From top to bottom are input images, corresponding ground truths, and prediction maps, respectively.}
\label{fig:trm}
\end{figure}

After that, we employ the max pooling operation to highlight camouflaged object regions and the average pooling operation to suppress background distractors. 
Following a summation operation and a global average pooling operation, we can obtain a one-dimensional weight map, which is then used to weigh the feature $f^{mm}$. 
Next, after two convolutional layers, we introduce an autonomous learnable parameter $\gamma$ as the weight to adaptively adjust the learned camouflaged feature to avoid omission or false detection. Finally, a residual mechanism is adopted to obtain the final camouflaged object feature ($f^{obj}$). It is formulated as: 
\begin{equation}
\left\{
\begin{aligned}
& f^{sum}  = {\rm Max}(f^{mm}) \oplus {\rm Avg}(f^{mm}), \\ 
& f^{w} = {\rm F}_{Conv3\times1}( {\rm F}_{Conv1\times3}( ({\rm GAP} (f^{sum}) \otimes f^{mm}) )), \\
& f^{obj} = f^{sin} \oplus \gamma \otimes f^{w},
\end{aligned}
\right.
\end{equation}
where ${\rm Max (\cdot)}$, ${\rm Avg (\cdot)}$, ${\rm GAP (\cdot)}$ denote max pooling, average pooling, and global average pooling operations, respectively. ${\rm F}_{Conva\times b}$ is a a$\times$b convolutional layer. $\oplus$ and $\otimes$ represent element-wise summation and element-wise multiplication, respectively. Note that $\gamma$ is initialized as 0 in our experiments.
Fig.~\ref{fig:trm} provides some predicted co-camouflage maps output from the proposed OFS module, showing the effectiveness of the proposed module in accurately discovering the camouflaged objects. 

\begin{figure}[tb]
\centering
\includegraphics[width=1\linewidth]{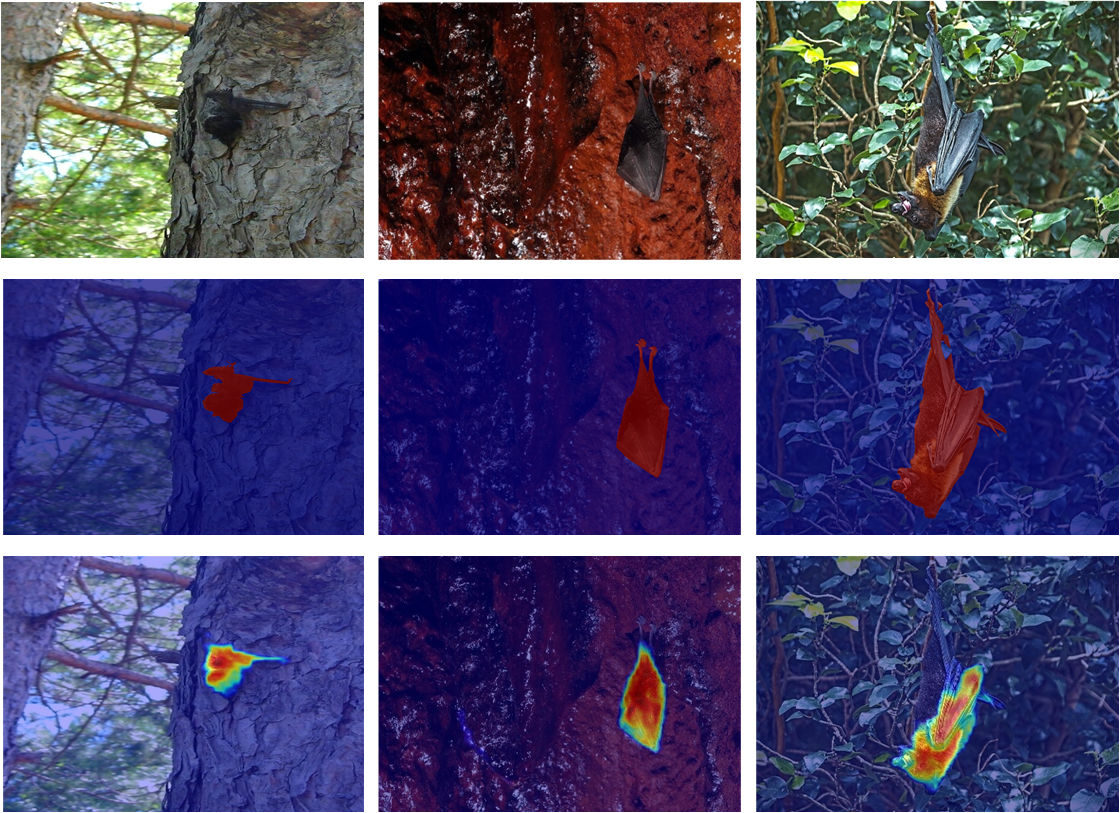} 
\caption{The effectiveness of our proposed model BBNet, which consists of backbone, CFE, OFS, and LGR modules. From top to bottom are input images, corresponding ground truths, and prediction maps, respectively.}
\label{fig:LGR}
\end{figure}

\subsection{Local-Global Refinement (LGR)} 
As shown in Fig.~\ref{fig:cfe} and Fig.~\ref{fig:trm}, the proposed CFE and OFS modules can accurately locate camouflaged objects from the inter-image branch and intra-image branch, respectively, but it is still difficult to accurately segment the camouflaged objects from their background due to the unclear boundary regions.    
Therefore, we first aggregate the inter-image feature ($f^{col}$) and intra-image feature ($f^{obj}$) by element-wise multiplication. Then, we design a local-global refinement module (LGR) to consummate the details of co-camouflaged objects. 

To explore local cues, inspired by YOLO~\cite{redmon2016you}, we first divide the aggregated feature ($f^{ag}$) into 3$\times$3 sub-blocks on the H$\times$W dimension, \textit{i.e.}, $f^{ag}_{i,j},~i,j=\{1,2,3\}$. Then, a Sigmoid function is applied to calculate the probability that each sub-block contains the common camouflaged objects. The sub-block with the highest probability is selected as local clues. After that, we adopt a convolutional layer with a kernel size of 1$\times$3 to strengthen the local features, and a convolution layer with a kernel of 3$\times$1 to restore the original size of the sub-block. The above operations can be denoted as: 
\begin{equation}
\left\{
\begin{aligned}
f_{max} & = {\rm F_{max}}(\sigma (f^{ag}_{i,j})), (i,j) = \{1,2,3\}, \\
f^{local} & = {\rm F}_{Conv3\times1}( {\rm F}_{Conv1\times3} (f_{max}) )
\end{aligned}
\right.
\end{equation}
where $\sigma (\cdot)$ denotes the Sigmoid function and ${\rm F_{max}}(\cdot)$ denotes selecting sub-block with maximum probability. 
Besides, for global feature refinement, we adopt the same convolution operation for the feature $f^{ag}$, which is formulated as: 
\begin{equation}
f^{global} = {\rm F}_{Conv3\times1}( {\rm F}_{Conv1\times 3} ( f^{ag} ) ),
\end{equation}
After that, we utilize the channel concatenation operation to fuse the local and global features, denoted as: 
\begin{equation}
f^{lgr} = {\rm F}_{Conv3\times3}( {\rm Cat}( \delta_\uparrow^3( f^{local}), f^{global}) ),
\end{equation}
where $\delta_\uparrow^3$ represents a 3$\times$ upsampling operation. 

To produce the final object mask prediction, we employ two simple decoders to integrate the refined co-camouflaged feature $f^{lgr}$ with the backbone feature $f_4$ and $f_3$, respectively. 
Specifically, we leverage a Sigmoid operation on feature $f^{lgr}$ as a weight to integrate with the backbone feature $f_4$ by element-wise multiplication, following an element-wise summation operation with $f_4$ and a 1$\times$1 convolution operation. 
Then, we apply the same operations to integrate the backbone feature $f_3$. Thus, we obtain the final co-camouflaged prediction maps. 
Fig.~\ref{fig:LGR} shows the effectiveness of the proposed LGR module, which provides more accurate object region and boundary predictions. As can be seen, the LGR module further explores and enhances the co-camouflaged feature details from local and global perspectives for co-camouflaged object prediction.

\subsection{Loss Function}
In this paper, we employ the weighted binary cross-entropy loss ($\mathcal{L}^{\omega}_{BCE}$) and weighted intersection over-union (IoU) loss ($\mathcal{L}^{\omega}_{IoU}$) to train our model. Note that the mask supervision is applied to the output of OFS and the final predictions of our model. The total loss is defined as:
\begin{equation}
\left \{
	\begin{aligned}
		\mathcal{L}_{OFS} &= \mathcal{L}^{\omega}_{BCE} (P^{OFS}, G)+ \mathcal{L}^{\omega}_{IoU} (P^{OFS}, G) \\
		\mathcal{L}_{total} &= \mathcal{L}^{\omega}_{BCE} (P, G)+ \mathcal{L}^{\omega}_{IoU} (P, G) +  \mathcal{L}_{OFS} \\	
	\end{aligned}
\right.
\end{equation}
where $G$ denotes the ground truth, and $P$ and $P^{OFS}$ denote the predictions of the proposed model and the proposed OFS, respectively. More details of these two losses can be found in~\cite{wei2020f3net,fan2022concealed}.

\begin{table*}[ht]
	\centering
	\caption{The quantitative comparison with 12 state-of-the-art models on CoCOD8K and its five sub-datasets under six widely used evaluation metrics, including S-measure ($S_{\alpha}$), MAE ($\mathcal{M}$), maximum F-measure ($F_{max}$), maximum E-measure ($E_{max}$), mean F-measure ($F_{mean}$), and mean E-measure ($E_{mean}$). $\uparrow$ means that the higher the score, the better. $\downarrow$ indicates that the lower the score, the better. Type: COD = camouflaged object detection models. \textcolor[rgb]{1,  0,  0}{Red} and \textcolor[rgb]{.439,  .678,  .278}{green} denote the best and the second-best results, respectively.}
	\renewcommand{\arraystretch}{1.15}	\renewcommand{\tabcolsep}{0.505mm}
	\begin{tabular}{c|c|cccccccccccc|l}
		\hline
		\multirow{4}[1]{*}{Datesets} & Method & SINet & JCOSOD & SINet-v2 & RankNet & C2FNet & PFNet & SegMaR & ERRNet & OCENet & BGNet & BgNet & BSANet  & \multirow{4}[1]{*}{Ours} \bigstrut[t]\\
		& Pub.   & CVPR  & CVPR  & PAMI  & CVPR  & IJCAI & CVPR & CVPR & PR & WACV & IJCAI & KBS & AAAI  &  \\
		& Year  & 2020 \cite{fan2020camouflaged}  & 2021 \cite{li2021uncertainty} & 2021 \cite{fan2022concealed}  & 2021 \cite{lv2021simultaneously}  & 2021 \cite{sun2021context} & 2021 \cite{mei2021camouflaged} & 2022 \cite{jia2022segment} & 2022 \cite{ji2022fast} & 2022 \cite{liu2022modeling} & 2022 \cite{sun2022boundary} & 2022 \cite{chen2022boundary} &2022 \cite{zhu2022can}  &  \\
		& Type  & COD   & COD   & COD   & COD   & COD   & COD & COD   & COD   & COD   & COD  & COD   & COD  &  \\ \hline
		\multirow{6}[1]{*}{CoCOD8K} & $S_{\alpha} \uparrow$     & 0.707 & 0.459 & 0.714 & 0.725 & 0.741 & 0.749 & 0.587 & 0.688 & 0.689 & 0.729 & 0.751 & \textcolor[rgb]{ .439,  .678,  .278}{0.755} & \textcolor[rgb]{ 1,  0,  0}{0.789} \\
		& $\mathcal{M} \downarrow$   & 0.058 & 0.258 & 0.065 & 0.065 & 0.059  & 0.055 & 0.098 & 0.063 & 0.059 & 0.053 & 0.052 & \textcolor[rgb]{ .439,  .678,  .278}{0.049} & \textcolor[rgb]{ 1,  0,  0}{0.046} \\
		& $E_{max}\uparrow$ & 0.835 & 0.623  & 0.816 & 0.822 & 0.830  & 0.830 & 0.705 & 0.804 & 0.801 & 0.828 & 0.845 & \textcolor[rgb]{ .439,  .678,  .278}{0.847} & \textcolor[rgb]{ 1,  0,  0}{0.855 } \\
		& $F_{max}\uparrow$ & 0.615 & 0.218 & 0.575 & 0.592 & 0.613 & 0.627 & 0.375 & 0.542 & 0.546 & 0.608 & 0.631 & \textcolor[rgb]{ .439,  .678,  .278}{0.638} & \textcolor[rgb]{ 1,  0,  0}{0.688} \\
		& $E_{mean}\uparrow$ & 0.730  & 0.411  & 0.729 & 0.743 & 0.778 & 0.781 & 0.621 & 0.715 & 0.720 & 0.764 & 0.781 & \textcolor[rgb]{ .439,  .678,  .278}{0.795} & \textcolor[rgb]{ 1,  0,  0}{0.840 } \\
		& $F_{mean}\uparrow$ & 0.567 & 0.134 & 0.525 & 0.540  & 0.577 & 0.592 & 0.318 & 0.502 & 0.516 & 0.577 & 0.591 & \textcolor[rgb]{ .439,  .678,  .278}{0.609} & \textcolor[rgb]{ 1,  0,  0}{0.670} \bigstrut[b]\\
		\hline
  
        \multirow{6}[1]{*}{Insect} & $S_{\alpha} \uparrow$     & 0.717 & 0.511 & 0.721 & 0.728 & 0.742 & 0.751 & 0.610 & 0.695 & 0.694 & 0.729 & 0.751 & \textcolor[rgb]{ .439,  .678,  .278}{0.754} & \textcolor[rgb]{ 1,  0,  0}{0.768} \\
		& $\mathcal{M} \downarrow$   & 0.058 & 0.236 & 0.066 & 0.067 & 0.062  & 0.057 & 0.094 & 0.065 & 0.060 & 0.055 & 0.055 & \textcolor[rgb]{ .439,  .678,  .278}{0.052} & \textcolor[rgb]{ 1,  0,  0}{0.048} \\
		& $E_{max}\uparrow$ & 0.834 & 0.631  & 0.821 & 0.819 & 0.828  & 0.826 & 0.730 & 0.805 & 0.803 & 0.825 & 0.841 & \textcolor[rgb]{ .439,  .678,  .278}{0.846} & \textcolor[rgb]{ 1,  0,  0}{0.849 } \\
		& $F_{max}\uparrow$ & 0.611 & 0.265 & 0.586 & 0.598 & 0.615 & 0.628 & 0.423 & 0.556 & 0.561 & 0.616 & 0.632 & \textcolor[rgb]{ .439,  .678,  .278}{0.639} & \textcolor[rgb]{ 1,  0,  0}{0.664} \\
		& $E_{mean}\uparrow$ & 0.726  & 0.422  & 0.732 & 0.734 & 0.777 & 0.778 & 0.641 & 0.720 & 0.721 & 0.759 & 0.778 & \textcolor[rgb]{ .439,  .678,  .278}{0.793} & \textcolor[rgb]{ 1,  0,  0}{0.794 } \\
		& $F_{mean}\uparrow$ & 0.553 & 0.158 & 0.532 & 0.544  & 0.581 & 0.594 & 0.365 & 0.513 & 0.526 & 0.582 & 0.593 & \textcolor[rgb]{ .439,  .678,  .278}{0.610} & \textcolor[rgb]{ 1,  0,  0}{0.631} \bigstrut[b]\\
		\hline
        \multirow{6}[1]{*}{Sea} & $S_{\alpha} \uparrow$     & 0.714 & 0.482 & 0.711 & 0.723 & 0.756 & 0.760 & 0.534 & 0.685 & 0.680 & 0.738 & 0.771 & \textcolor[rgb]{ .439,  .678,  .278}{0.770} & \textcolor[rgb]{ 1,  0,  0}{0.788} \\
		& $\mathcal{M} \downarrow$   & 0.092 & 0.299 & 0.097 & 0.094 & 0.080  & 0.076 & 0.144 & 0.096 & 0.097 & 0.080 & 0.072 & \textcolor[rgb]{ .439,  .678,  .278}{0.069} & \textcolor[rgb]{ 1,  0,  0}{0.065} \\
		& $E_{max}\uparrow$ & 0.830 & 0.595  & 0.807 & 0.813 & 0.831  & 0.837 & 0.695 & 0.799 & 0.796 & 0.833 & \textcolor[rgb]{ .439,  .678,  .278}{0.850} & \textcolor[rgb]{ .439,  .678,  .278}{0.850} & \textcolor[rgb]{ 1,  0,  0}{0.857 } \\
		& $F_{max}\uparrow$ & 0.664 & 0.279 & 0.626 & 0.613 & 0.680 & 0.689 & 0.451 & 0.599 & 0.599 & 0.672 & 0.699 & \textcolor[rgb]{ .439,  .678,  .278}{0.701} & \textcolor[rgb]{ 1,  0,  0}{0.726} \\
		& $E_{mean}\uparrow$ & 0.731  & 0.399  & 0.729 & 0.748 & 0.793 & 0.797 & 0.581 & 0.711 & 0.713 & 0.777 & 0.804 & \textcolor[rgb]{ .439,  .678,  .278}{0.810} & \textcolor[rgb]{ 1,  0,  0}{0.821 } \\
		& $F_{mean}\uparrow$ & 0.608 & 0.167 & 0.578 & 0.599  & 0.649 & 0.661 & 0.351 & 0.557 & 0.564 & 0.644 & 0.667 & \textcolor[rgb]{ .439,  .678,  .278}{0.677} & \textcolor[rgb]{ 1,  0,  0}{0.701} \bigstrut[b]\\
		\hline
         \multirow{6}[1]{*}{Fly} & $S_{\alpha} \uparrow$     & 0.787 & 0.473 & 0.762 & 0.769 & 0.789 & 0.794 & 0.594 & 0.734 & 0.746 & 0.780 & 0.797 & \textcolor[rgb]{ .439,  .678,  .278}{0.807} & \textcolor[rgb]{ 1,  0,  0}{0.823} \\
		& $\mathcal{M} \downarrow$   & 0.034 & 0.267 & 0.042 & 0.043 & 0.038  & 0.036 & 0.084 & 0.040 & 0.035 & 0.034 & 0.033 & \textcolor[rgb]{ .439,  .678,  .278}{0.030} & \textcolor[rgb]{ 1,  0,  0}{0.026} \\
		& $E_{max}\uparrow$ & 0.872 & 0.636  & 0.859 & 0.866 & 0.864  & 0.867 & 0.722 & 0.842 & 0.840 & 0.766 & \textcolor[rgb]{ .439,  .678,  .278}{0.878} & \textcolor[rgb]{ .439,  .678,  .278}{0.878} & \textcolor[rgb]{ 1,  0,  0}{0.881 } \\
		& $F_{max}\uparrow$ & 0.679 & 0.174 & 0.636 & 0.649 & 0.669 & 0.684 & 0.367 & 0.591 & 0.598 & 0.655 & 0.688 & \textcolor[rgb]{ .439,  .678,  .278}{0.701} & \textcolor[rgb]{ 1,  0,  0}{0.734} \\
		& $E_{mean}\uparrow$ & 0.800  & 0.405  & 0.769 & 0.745 & 0.814 & 0.822 & 0.631 & 0.751 & 0.769 & 0.809 & 0.837 & \textcolor[rgb]{ .439,  .678,  .278}{0.838} & \textcolor[rgb]{ 1,  0,  0}{0.840 } \\
		& $F_{mean}\uparrow$ & 0.634 & 0.108 & 0.581 & 0.592  & 0.625 & 0.641 & 0.311 & 0.552 & 0.573 & 0.632 & 0.573 & \textcolor[rgb]{ .439,  .678,  .278}{0.662} & \textcolor[rgb]{ 1,  0,  0}{0.690} \bigstrut[b]\\
		\hline
         \multirow{6}[1]{*}{Amohibious} & $S_{\alpha} \uparrow$     & 0.766 & 0.509 & 0.765 & 0.789 & 0.804 & 0.813 & 0.610 & 0.746 & 0.746 & 0.777 & 0.812 & \textcolor[rgb]{ .439,  .678,  .278}{0.816} & \textcolor[rgb]{ 1,  0,  0}{0.830} \\
		& $\mathcal{M} \downarrow$   & 0.059 & 0.269 & 0.066 & 0.062 & 0.056  & 0.051 & 0.110 & 0.063 & 0.062 & 0.055 & 0.047 & \textcolor[rgb]{ .439,  .678,  .278}{0.045} & \textcolor[rgb]{ 1,  0,  0}{0.041} \\
		& $E_{max}\uparrow$ & 0.867 & 0.632  & 0.866 & 0.866 & 0.885  & 0.884 & 0.728 & 0.865 & 0.834 & 0.852 & \textcolor[rgb]{ .439,  .678,  .278}{0.889} & \textcolor[rgb]{ .439,  .678,  .278}{0.892} & \textcolor[rgb]{ 1,  0,  0}{0.895 } \\
		& $F_{max}\uparrow$ & 0.694 & 0.309 & 0.671 & 0.699 & 0.727 & 0.734 & 0.465 & 0.651 & 0.647 & 0.691 & 0.735 & \textcolor[rgb]{ .439,  .678,  .278}{0.742} & \textcolor[rgb]{ 1,  0,  0}{0.762} \\
		& $E_{mean}\uparrow$ & 0.774  & 0.411  & 0.777 & 0.803 & 0.844 & 0.845 & 0.640 & 0.781 & 0.770 & 0.808 & 0.845 & \textcolor[rgb]{ .439,  .678,  .278}{0.848} & \textcolor[rgb]{ 1,  0,  0}{0.858 } \\
		& $F_{mean}\uparrow$ & 0.655 & 0.179 & 0.624 & 0.657  & 0.692 & 0.702 & 0.395 & 0.621 & 0.625 & 0.667 & 0.700 & \textcolor[rgb]{ .439,  .678,  .278}{0.717} & \textcolor[rgb]{ 1,  0,  0}{0.737} \bigstrut[b]\\
		\hline
         \multirow{6}[1]{*}{Land} & $S_{\alpha} \uparrow$     & 0.703 & 0.470 & 0.694 & 0.714 & 0.730 & 0.734 & 0.563 & 0.673 & 0.677 & 0.710 & 0.737 & \textcolor[rgb]{ .439,  .678,  .278}{0.747} & \textcolor[rgb]{ 1,  0,  0}{0.765} \\
		& $\mathcal{M} \downarrow$   & 0.059 & 0.270 & 0.067 & 0.067 & 0.059  & 0.057 & 0.123 & 0.066 & 0.060 & 0.056 & 0.054 & \textcolor[rgb]{ .439,  .678,  .278}{0.051} & \textcolor[rgb]{ 1,  0,  0}{0.047} \\
		& $E_{max}\uparrow$ & 0.818 & 0.599  & 0.797 & 0.814 & 0.823  & 0.822 & 0.683 & 0.786 & 0.789 & 0.811 & \textcolor[rgb]{ .439,  .678,  .278}{0.837} & \textcolor[rgb]{ .439,  .678,  .278}{0.838} & \textcolor[rgb]{ 1,  0,  0}{0.850 } \\
		& $F_{max}\uparrow$ & 0.590 & 0.183 & 0.553 & 0.578 & 0.596 & 0.610 & 0.334 & 0.523 & 0.530 & 0.578 & 0.611 & \textcolor[rgb]{ .439,  .678,  .278}{0.628} & \textcolor[rgb]{ 1,  0,  0}{0.656} \\
		& $E_{mean}\uparrow$ & 0.718  & 0.402  & 0.710 & 0.729 & 0.767 & 0.765 & 0.600 & 0.701 & 0.703 & 0.741 & 0.767 & \textcolor[rgb]{ .439,  .678,  .278}{0.783} & \textcolor[rgb]{ 1,  0,  0}{0.795 } \\
		& $F_{mean}\uparrow$ & 0.540 & 0.115 & 0.504 & 0.527  & 0.562 & 0.577 & 0.277 & 0.487 & 0.500 & 0.849 & 0.573 & \textcolor[rgb]{ .439,  .678,  .278}{0.596} & \textcolor[rgb]{ 1,  0,  0}{0.626} \bigstrut[b]\\
		\hline
	\end{tabular}%
	\label{tab:quan_cod}%
\end{table*}%

\begin{figure*}[ht]
	\includegraphics[width=1.0\linewidth]{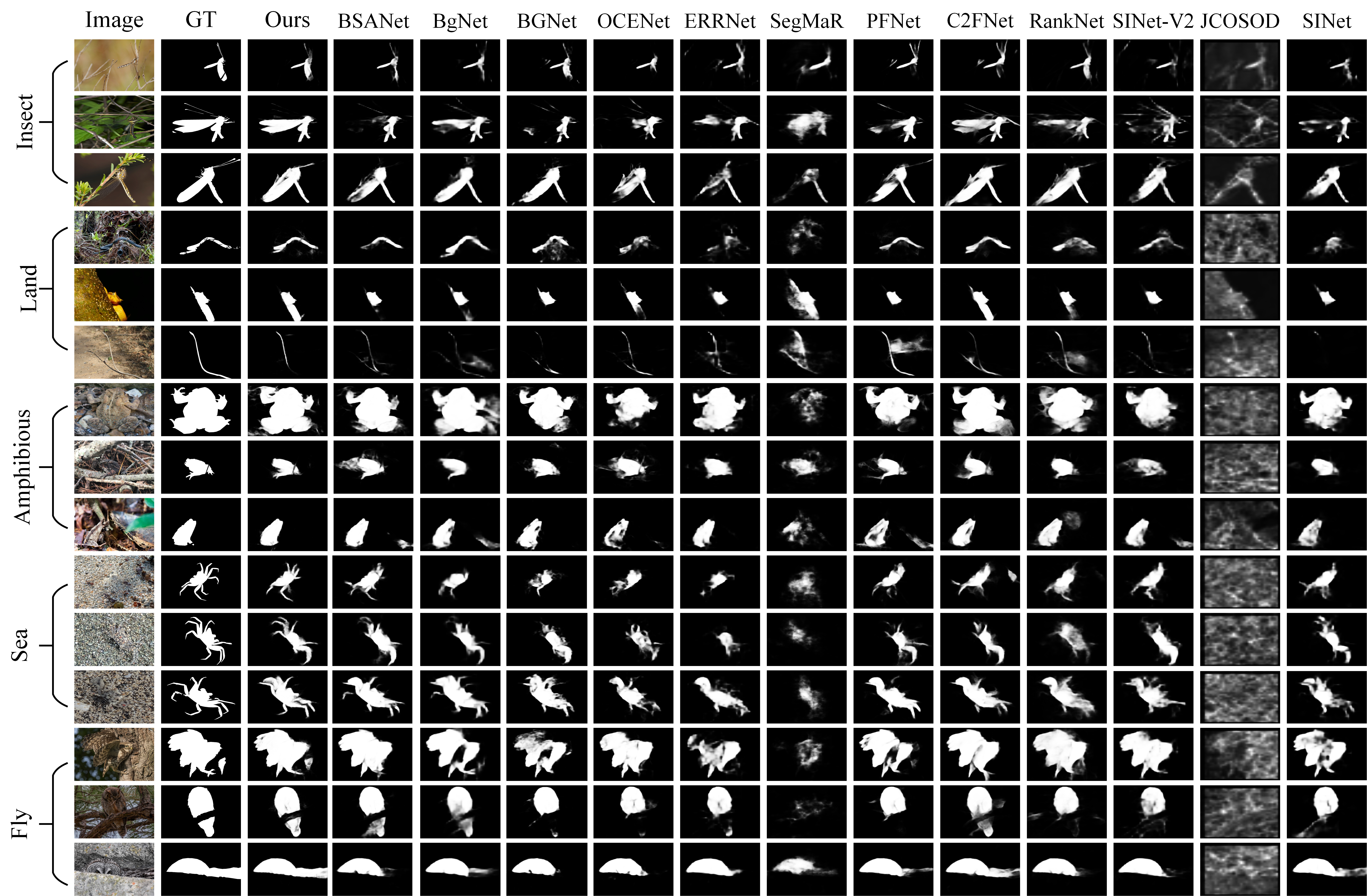}
	\caption{Visual comparison of the proposed model with other 12 existing state-of-the-art COD algorithms on on five sub-datasets of the proposed CoCOD8K datasets. The first and the second columns denote the input images and ground-truths, respectively. The third one represents the predicted maps of our model. Please zoom in for more details.}
	\label{fig:viscom_cod} 
\end{figure*}

\section{Experiments}
\label{sec:experiment}
This section conducts a comprehensive experimental evaluation of the proposed BBNet on our proposed CoCOD8K dataset. Furthermore, we carry out multiple ablation experiments to verify the effectiveness of each proposed module.

\subsection{Experiment Setup}
The proposed BBNet is implemented with the PyTorch toolbox on an NVIDIA 1080 Ti GPU.  The Res2Net~\cite{gao2019res2net} is adopted as our backbone. All the experiments are conducted on our proposed CoCOD8K dataset and the input image is resized to 288$ \times$288. During the training stage, the momentum and weight decay are set to 0.9 and 0.0005, respectively. The learning rate is 1e-4, and the batch size is 10.

\begin{figure*}[ht]
	\centering
	\includegraphics[width=0.98\linewidth]{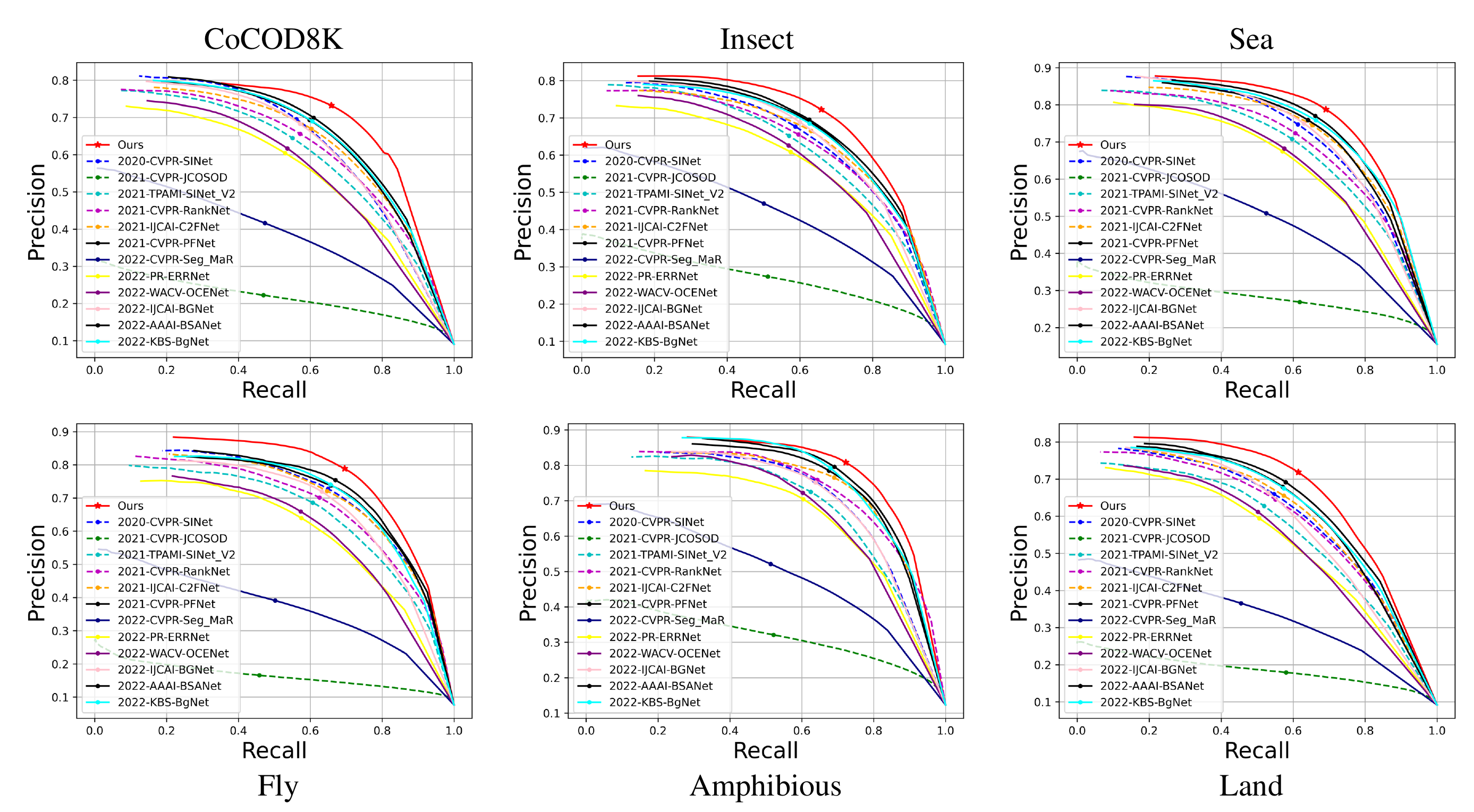}
	\caption{PR curves of the proposed model and other 12 SOTA COD methods on CoCOD8K and its 5 sub-datasets.}
	\label{fig:pr_cod}       
\end{figure*}

\begin{table*}[ht]
	\centering
	\caption{The quantitative comparison with 6 state-of-the-art CoSOD models on CoCOD8K and its five sub-datasets under six widely used evaluation metrics. Type: CoSOD = collaborative salient object detection models. \textcolor[rgb]{ 1,  0,  0}{Red} and \textcolor[rgb]{ .439,  .678,  .278}{green} denote the best and the second-best results, respectively.}
	\renewcommand{\arraystretch}{1.15}	\renewcommand{\tabcolsep}{3.9mm}
	\begin{tabular}{c|c|cccccc|l}
		\hline
		\multirow{4}[1]{*}{Datesets} & Method & \multicolumn{1}{c}{GICD\cite{zhang2020gradient}}  & ICNet\cite{jin2020icnet} & CoEGNet\cite{fan2021re} & CADC\cite{zhang2021summarize}  & GCONet\cite{fan2021group} & DCFM \cite{yu2022democracy} & \multirow{4}[1]{*}{Ours} \bigstrut[t]\\
		& Pub   & \multicolumn{1}{c}{ECCV} & NeurIPS & TPAMI   & ICCV & CVPR  & CVPR & \\
		& Year  & \multicolumn{1}{c}{2020} & 2020  & 2021  & 2021  & 2021  & 2022 & \\
		& Type  & \multicolumn{1}{c}{CoSOD} & CoSOD & CoSOD & CoSOD & CoSOD & CoSOD & \\ \hline
		\multirow{6}[1]{*}{CoCOD8K} & $S_{\alpha} \uparrow$     & 0.657 & 0.718 & 0.530  & 0.695 & 0.687 &\textcolor[rgb]{.439,  .678,  .278}{0.728} & \textcolor[rgb]{ 1,  0,  0}{0.789} \\
		& $\mathcal{M} \downarrow$    & \multicolumn{1}{c}{0.072} & 0.076 & 0.168 & 0.065 & 0.066 &  \textcolor[rgb]{ .439,  .678,  .278}{0.062} & \textcolor[rgb]{ 1,  0,  0}{0.046} \\
		& $E_{max}\uparrow$ & 0.733 & \textcolor[rgb]{ .439,  .678,  .278}{0.835} & 0.643 & 0.822 & 0.795 & 0.815 & \textcolor[rgb]{ 1,  0,  0}{0.855} \\
		& $F_{max}\uparrow$ & 0.525 & \textcolor[rgb]{.439,  .678,  .278}{0.621} & 0.286 &0.553 & 0.548 & 0.590 & \textcolor[rgb]{ 1,  0,  0}{0.688} \\
		& $E_{mean}\uparrow$ & 0.725 & \textcolor[rgb]{ .439,  .678,  .278}{0.817} & 0.579 & 0.761 & 0.739 & 0.750 & \textcolor[rgb]{ 1,  0,  0}{0.840 } \\
		& $F_{mean}\uparrow$ & 0.521 & \textcolor[rgb]{ .439,  .678,  .278}{0.595} & 0.262 & 0.519 & 0.522 &0.548 & \textcolor[rgb]{ 1,  0,  0}{0.670} \bigstrut[b]\\
		\hline
        \multirow{6}[1]{*}{Insect} & $S_{\alpha} \uparrow$     & 0.617 & 0.728 & 0.696  & 0.578 & 0.689 &\textcolor[rgb]{.439,  .678,  .278}{0.730} & \textcolor[rgb]{ 1,  0,  0}{0.768} \\
		& $\mathcal{M} \downarrow$    & 0.074 & 0.076 & 0.067 & 0.100 & 0.067 &  \textcolor[rgb]{ .439,  .678,  .278}{0.062} & \textcolor[rgb]{ 1,  0,  0}{0.048} \\
		& $E_{max}\uparrow$ & 0.681 & \textcolor[rgb]{ .439,  .678,  .278}{0.840} & 0.826 & 0.710 & 0.797 & 0.820 & \textcolor[rgb]{ 1,  0,  0}{0.846} \\
		& $F_{max}\uparrow$ & 0.475 & \textcolor[rgb]{.439,  .678,  .278}{0.628} & 0.562 & 0.423 & 0.559 & 0.595 & \textcolor[rgb]{ 1,  0,  0}{0.664} \\
		& $E_{mean}\uparrow$ & 0.668 & \textcolor[rgb]{ .439,  .678,  .278}{0.823} & 0.757 & 0.692 & 0.732 & 0.750 & \textcolor[rgb]{ 1,  0,  0}{0.794 } \\
		& $F_{mean}\uparrow$ & 0.470 & \textcolor[rgb]{ .439,  .678,  .278}{0.608} & 0.525 & 0.408 & 0.526 &0.553 & \textcolor[rgb]{ 1,  0,  0}{0.631} \bigstrut[b]\\
		\hline
        \multirow{6}[1]{*}{Sea} & $S_{\alpha} \uparrow$     & 0.573 & 0.722 & 0.688  & 0.546 & 0.655 &\textcolor[rgb]{.439,  .678,  .278}{0.737} & \textcolor[rgb]{ 1,  0,  0}{0.788} \\
		& $\mathcal{M} \downarrow$    & 0.141 & 0.103 & 0.099 & 0.151 & 0.110 &  \textcolor[rgb]{ .439,  .678,  .278}{0.091} & \textcolor[rgb]{ 1,  0,  0}{0.065} \\
		& $E_{max}\uparrow$ & 0.651 & \textcolor[rgb]{ .439,  .678,  .278}{0.815} & 0.812 & 0.655 & 0.763 & 0.807 & \textcolor[rgb]{ 1,  0,  0}{0.857} \\
		& $F_{max}\uparrow$ & 0.435 & \textcolor[rgb]{.439,  .678,  .278}{0.661} & 0.594 & 0.418 & 0.575 & 0.649 & \textcolor[rgb]{ 1,  0,  0}{0.726} \\
		& $E_{mean}\uparrow$ & 0.645 & \textcolor[rgb]{ .439,  .678,  .278}{0.800} & 0.768 & 0.613 & 0.703 & 0.763 & \textcolor[rgb]{ 1,  0,  0}{0.821 } \\
		& $F_{mean}\uparrow$ & 0.433 & \textcolor[rgb]{ .439,  .678,  .278}{0.644} & 0.568 & 0.371 & 0.516 &0.617 & \textcolor[rgb]{ 1,  0,  0}{0.701} \bigstrut[b]\\
		\hline
        \multirow{6}[1]{*}{Fly} & $S_{\alpha} \uparrow$     & 0.786 & 0.760 & 0.751  & 0.578 & 0.748 &\textcolor[rgb]{.439,  .678,  .278}{0.775} & \textcolor[rgb]{ 1,  0,  0}{0.823} \\
		& $\mathcal{M} \downarrow$    & 0.035 & 0.054 & 0.040 & 0.104 & 0.042 &  \textcolor[rgb]{ .439,  .678,  .278}{0.040} & \textcolor[rgb]{ 1,  0,  0}{0.026} \\
		& $E_{max}\uparrow$ & 0.862 & \textcolor[rgb]{ .439,  .678,  .278}{0.879} & 0.867 & 0.678 & 0.838 & 0.842 & \textcolor[rgb]{ 1,  0,  0}{0.881} \\
		& $F_{max}\uparrow$ & 0.721 & \textcolor[rgb]{.439,  .678,  .278}{0.690} & 0.627 & 0.350 & 0.616 & 0.645 & \textcolor[rgb]{ 1,  0,  0}{0.734} \\
		& $E_{mean}\uparrow$ & 0.857 & \textcolor[rgb]{ .439,  .678,  .278}{0.806} & 0.805 & 0.656 & 0.798 & 0.785 & \textcolor[rgb]{ 1,  0,  0}{0.840 } \\
		& $F_{mean}\uparrow$ & 0.715 & \textcolor[rgb]{ .439,  .678,  .278}{0.639} & 0.587 & 0.340 & 0.588 &0.599 & \textcolor[rgb]{ 1,  0,  0}{0.690} \bigstrut[b]\\
		\hline
        \multirow{6}[1]{*}{Amphibious} & $S_{\alpha} \uparrow$     & 0.739 & 0.778 & 0.747  & 0.591 & 0.731 &\textcolor[rgb]{.439,  .678,  .278}{0.797} & \textcolor[rgb]{ 1,  0,  0}{0.830} \\
		& $\mathcal{M} \downarrow$    & 0.069 & 0.074 & 0.069 & 0.119 & 0.071 &  \textcolor[rgb]{ .439,  .678,  .278}{0.058} & \textcolor[rgb]{ 1,  0,  0}{0.041} \\
		& $E_{max}\uparrow$ & 0.826 & \textcolor[rgb]{ .439,  .678,  .278}{0.872} & 0.851 & 0.687 & 0.833 & 0.881 & \textcolor[rgb]{ 1,  0,  0}{0.895} \\
		& $F_{max}\uparrow$ & 0.684 & \textcolor[rgb]{.439,  .678,  .278}{0.724} & 0.646 & 0.453 & 0.637 & 0.714 & \textcolor[rgb]{ 1,  0,  0}{0.762} \\
		& $E_{mean}\uparrow$ & 0.819 & \textcolor[rgb]{ .439,  .678,  .278}{0.855} & 0.801 & 0.665 & 0.788 & 0.824 & \textcolor[rgb]{ 1,  0,  0}{0.858 } \\
		& $F_{mean}\uparrow$ & 0.681 & \textcolor[rgb]{ .439,  .678,  .278}{0.689} & 0.614 & 0.432 & 0.617 &0.670 & \textcolor[rgb]{ 1,  0,  0}{0.737} \bigstrut[b]\\
		\hline
        \multirow{6}[1]{*}{Land} & $S_{\alpha} \uparrow$     & 0.720 & 0.702 & 0.695  & 0.549 & 0.672 &\textcolor[rgb]{.439,  .678,  .278}{0.715} & \textcolor[rgb]{ 1,  0,  0}{0.765} \\
		& $\mathcal{M} \downarrow$    & 0.057 & 0.077 & 0.061 & 0.116 & 0.068 &  \textcolor[rgb]{ .439,  .678,  .278}{0.062} & \textcolor[rgb]{ 1,  0,  0}{0.047} \\
		& $E_{max}\uparrow$ & 0.797 & \textcolor[rgb]{ .439,  .678,  .278}{0.823} & 0.818 & 0.650 & 0.783 & 0.808 & \textcolor[rgb]{ 1,  0,  0}{0.850} \\
		& $F_{max}\uparrow$ & 0.626 & \textcolor[rgb]{.439,  .678,  .278}{0.603} & 0.555 & 0.320 & 0.529 & 0.575 & \textcolor[rgb]{ 1,  0,  0}{0.656} \\
		& $E_{mean}\uparrow$ & 0.792 & \textcolor[rgb]{ .439,  .678,  .278}{0.806} & 0.751 & 0.633 & 0.719 & 0.737 & \textcolor[rgb]{ 1,  0,  0}{0.795 } \\
		& $F_{mean}\uparrow$ & 0.622 & \textcolor[rgb]{ .439,  .678,  .278}{0.580} & 0.522 & 0.308 & 0.499 &0.535 & \textcolor[rgb]{ 1,  0,  0}{0.626} \bigstrut[b]\\
		\hline
	\end{tabular}%
	\label{tab:quan_cosod}%
\end{table*}%

\begin{figure*}[ht]
    \centering
	\includegraphics[width=1.0\linewidth]{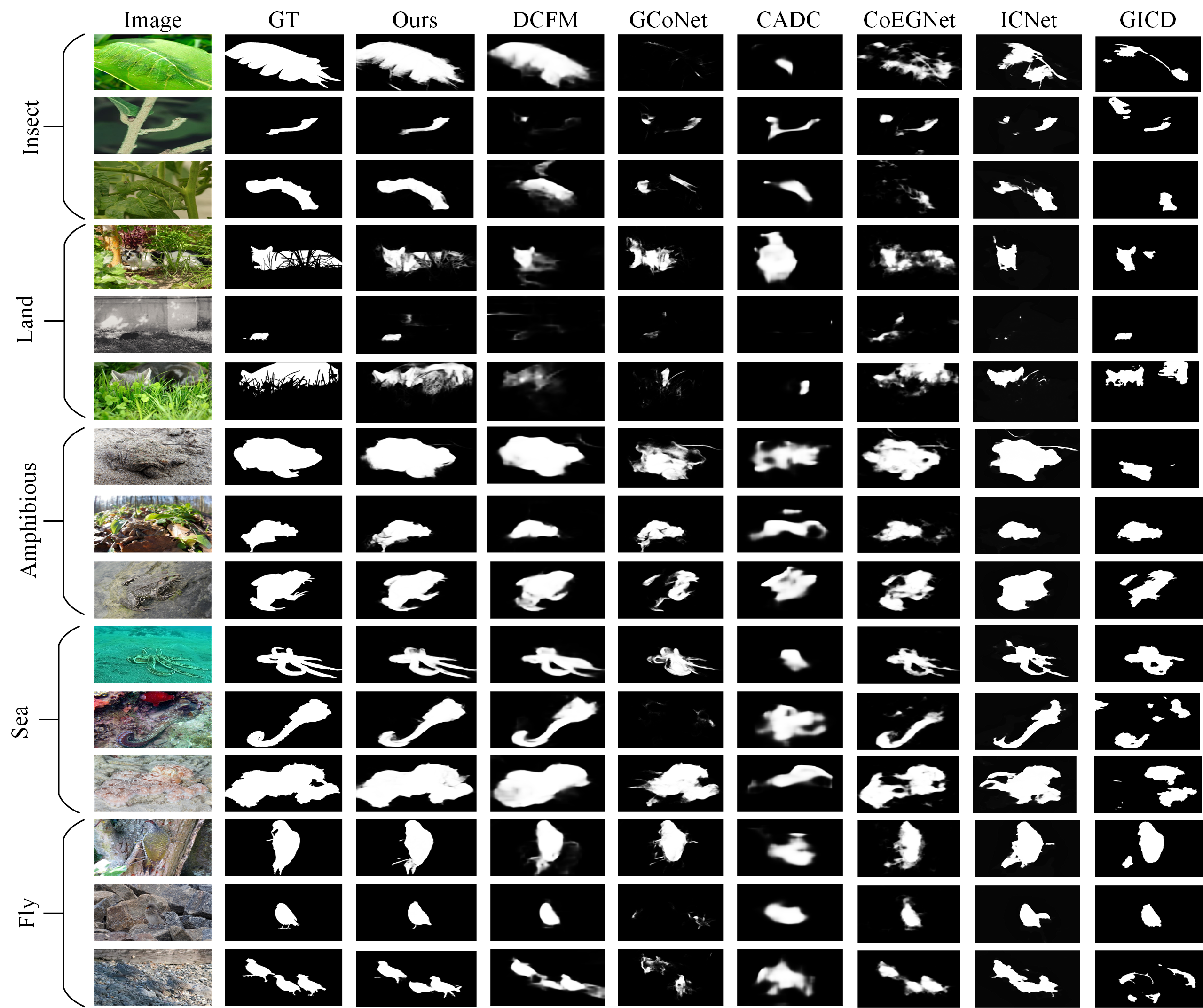}
	\caption{Visual comparison of the proposed model with other 6 existing state-of-the-art CoSOD methods on five sub-datasets of the proposed CoCOD8K datasets. The first and the second rows denote the input images and ground-truths, respectively. The third row represents the predicted maps of our model. Please zoom in for more details. }
	\label{fig:viscomp_cosod}       
\end{figure*}

\begin{figure*}[ht]
	\centering
	\includegraphics[width=0.95\linewidth]{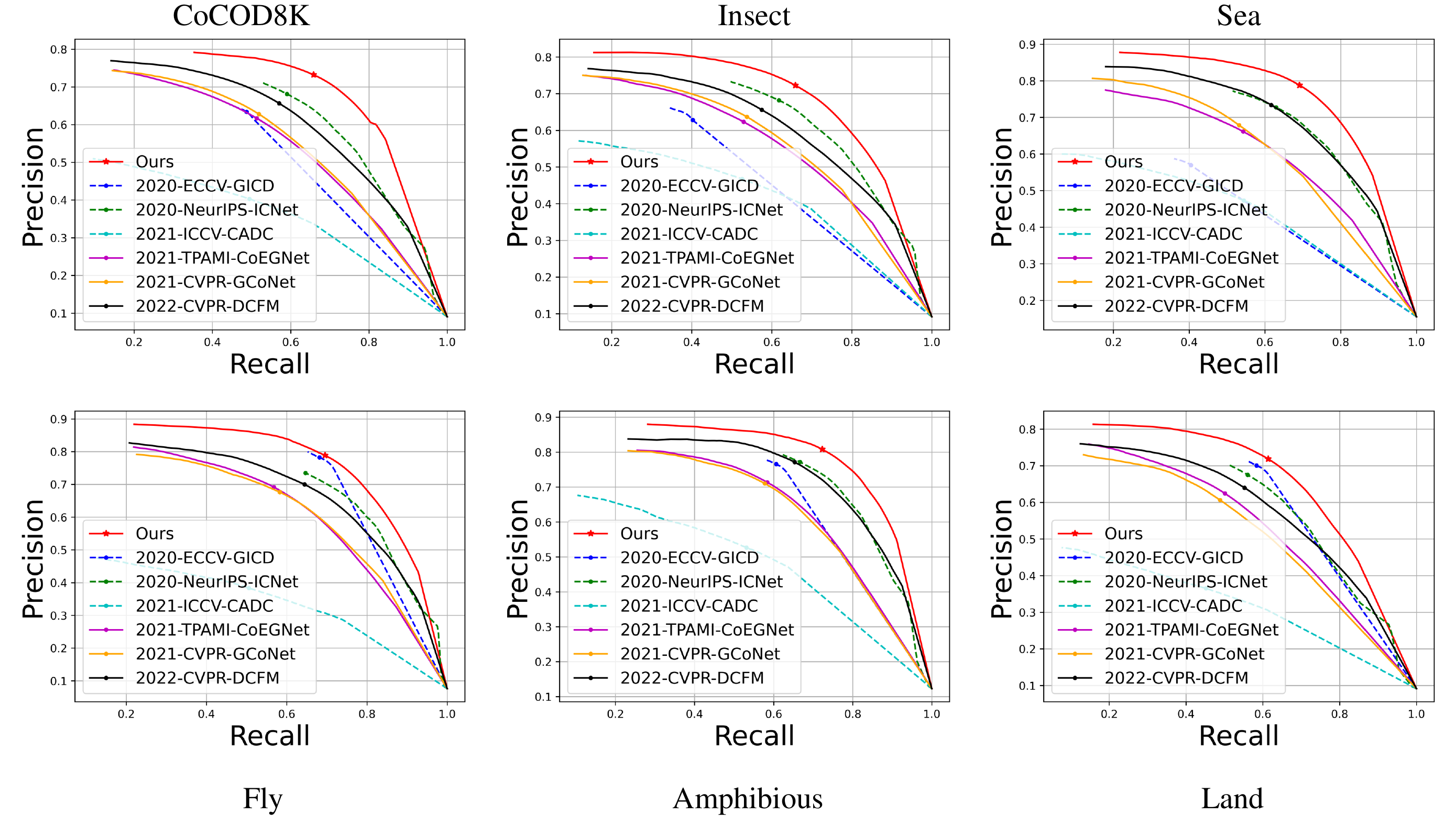}
	\caption{PR curves of the proposed method and other 6 state-of-the-art CoSOD methods on CoCOD8K and its 5 sub-datasets.}
	\label{fig:pr_cosod}       
\end{figure*}

\subsection{Evaluation metrics} 
To evaluate the performance of the proposed method, five widely used metrics are employed, including mean absolute error (MAE)~\cite{cheng2013efficient}, S-measure~\cite{fan2017structure_measure}, F-measure~\cite{achanta2009frequency-tuned}, E-measure~\cite{fan2018enhanced-alignment} and precision-recall (PR) curve.

\begin{itemize}
    \item \textbf{MAE} ($\mathcal{M}$) is a simple evaluation metric that measures the absolute difference between a predicted map and ground truth in a pixel-by-pixel manner. The larger the error, the larger the MAE. 
    
    \item \textbf{F-measure} is designed to evaluate the weighted harmonic mean of precision and recall. The co-camouflage maps are first binarized by thresholding pixels with a series of fixed integers from 0 to 255, where each threshold corresponds to a binary prediction. The predicted and ground-truth binary maps are compared to get precision and recall values. $F_\beta$ is typically chosen as the F-measure score that corresponds to the best-fixed threshold for the whole dataset. Note that we report the mean F-measure ($F_{mean}$) and max F-measure ($F_{max}$) in the experiments. 
    
    \item \textbf{E-measure} is a perceptual metric that evaluates both local pixel-level matching and global image-level statistics between the predicted map and ground truth. Note that we report the mean E-measure ($E_{mean}$) and max E-measure ($E_{max}$) in the experiments.
    
    \item \textbf{S-measure} ($S_\alpha$) is an evaluation metric to measure the structural similarity between a predicted map and the corresponding ground truth, which considers the region-oriented structural similarity and object-oriented structural similarity at the same time.
    
    \item \textbf{PR curve} shows the trade-off between precision and recall for different thresholds. The higher the area under the curve, the higher the recall and precision; that is, the better the algorithm performance.
\end{itemize}

\subsection{Comparison with the State-of-the-arts} \label{sec:comparison}
In this section, we made a detailed comparison between our proposed model and the other representative models. Collaborative camouflaged object detection (CoCOD) aims to simultaneously detect camouflaged objects with the same attributes from multiple images. To the best of our knowledge, research on CoCOD has not been publicly reported. As a result, we compare our proposed model with 12 recently proposed representative camouflaged object detection models, including SINet \cite{fan2020camouflaged}, JCOSOD \cite{li2021uncertainty}, SINet-v2 \cite{fan2022concealed}, RankNet \cite{lv2021simultaneously}, C2FNet \cite{sun2021context}, PFNet \cite{mei2021camouflaged}, SegMaR \cite{jia2022segment}, ERRNet \cite{ji2022fast}, OCENet \cite{liu2022modeling}, BGNet \cite{sun2022boundary} BgNet \cite{chen2022boundary}, and BSANet \cite{zhu2022can}.  
Furthermore, camouflaged objects can also be regarded as special salient objects in some low-contrast situations. Thus, we also compare the proposed method with 6 recently published cutting-edge deep CoSOD algorithms, \textit{i.e.}, GICD \cite{zhang2020gradient}, ICNet \cite{jin2020icnet}, CoEGNet \cite{fan2021re}, CADC \cite{zhang2021summarize}, GCoNet \cite{fan2021group}, and DCFM \cite{yu2022democracy}. 
It is worth noting that the results of all these methods are generated by models that are retrained on our CoCOD8K dataset based on the code released by the authors. 

\subsubsection{Quantitative Comparison with COD Models on CoCOD8K}
The detailed quantitative experimental results are shown in Tab.~\ref{tab:quan_cod}. It can be seen that the proposed method outperforms the other 12 competitors and achieves the new state-of-the-art (SOTA) performance on CoCOD8K and its 5 sub-datasets under six evaluation metrics.
Specifically, on the CoCOD8K dataset, compared with the second-best model BSANet, the performance gains are 4.50\% (0.034), 5.66\% (0.045), 9.03\% (0.055), 0.94\% (0.008), 7.05\% (0.045), 6.12\% (0.003) in terms of $S_\alpha$, $E_{mean}$, $F_{mean}$, $E_{max}$, $F_{max}$, and $\mathcal{M}$, respectively.

Among these five sub-datasets, the amphibious sub-dataset has the smallest number of image groups in the training set (only $\sim$8.23\%). 
This means the network can only capture a few clues about such co-camouflaged objects. Besides, the shape and appearance of amphibious creatures are extremely similar to their surroundings, which greatly increases the difficulty of accurate co-camouflaged object detection. Nonetheless, our method achieves superior performance, demonstrating its effectiveness for co-camouflaged object detection. In particular, $\mathcal{M}$ score decreases by 8.89\% (0.004) and $F_{mean}$ score increases by 2.79\% (0.020). 
The land sub-dataset contains a total of 21 species of organisms, which is the most extensive coverage among the five sub-datasets. In addition, the land sub-dataset also has the largest number of images in the training set ($\sim$27.50\%), which means the network is more likely to be well-trained to excavate effective features for accurate co-camouflaged object detection. Specifically, $\mathcal{M}$ score decreases by 7.84\% (0.004) and $F_{mean}$ score increases by 5.03\% (0.030) compared with the second-best model BSANet.

\subsubsection{Visual Comparison with COD Models on CoCOD8K} 
Fig.~\ref{fig:viscom_cod} provides the visual comparison of our proposed method with other COD competitors, which intuitively illustrates the outstanding performance of our proposed model. We can see that our proposed method provides more complete object regions, clear object boundaries, and fewer background disturbances compared with other methods. For example, in the first image group of Fig.~\ref{fig:viscom_cod}, our method can discover and segment the dragonfly object completely, while other methods either provide incomplete object predictions or introduce background distractors. Another example can be seen in the land group of Fig.~\ref{fig:viscom_cod}. The proposed method predicts the objects that are much closer to ground-truths, while other competitors only partially detect the snakes or do not detect these objects. 
In summary, our model can not only handle multiple complex environments simultaneously but also provide high-accuracy predicted maps. In addition, Fig.~\ref{fig:pr_cod} shows the PR curves of the proposed method and other 12 COD methods on CoCOD8K and its 5 sub-datasets. It can be observed that the proposed method outperforms other COD methods, where all the curves of our method are above those generated by other compared methods.

\subsubsection{Quantitative Comparison with CoSOD models on CoCOD8K}  
Tab.~\ref{tab:quan_cosod} shows the quantitative comparison of the proposed model and 6 state-of-the-art CoSOD methods. We can see that the proposed method achieves superior performance to other competitors under six widely-used evaluation metrics on CoCOD8K and its sub-datasets.
Specifically, on the CoCOD8K dataset, our proposed method achieves significant performance improvements by 9.89\% (0.071), 2.82\% (0.023), 11.60\% (0.069), 2.40\% (0.020), 9.98\% (0.062) and 39.47\% (0.03) in terms of $S_\alpha$, $E_{mean}$, $F_{mean}$, $E_{max}$, $F_{max}$, and $\mathcal{M}$, respectively, compared with the second-best method ICNet.
For 5 sub-datasets, our proposed method achieves different degrees of performance improvement on the six metrics. Particularly,  compared with the recently proposed DCFM method, our method greatly drops $\mathcal{M}$ by 22.58\%, 28.57\%, 35.00\%, 29.31\%, and 24.19\% on Insect, Sea, Fly, Amphibious, and Land sub-datasets, respectively.

\subsubsection{Visual Comparison with CoSOD models on CoCOD8K} 
Fig.~\ref{fig:viscomp_cosod} exhibits visual comparisons of the proposed model with other compared CoSOD methods on the CoCOD8K dataset and its 5 sub-datasets.  
We enumerate several representative challenging co-camouflaged image groups of various creature categories, including worm, cat, toad, octopus, and bird. It can be seen that other CoSOD models either can not completely segment the co-camouflaged objects or probably mistake noise or other interference as co-camouflaged objects, especially in the bird and worm groups, where leaves or stones are wrongly recognized as camouflaged objects. Intuitively, our method can locate and segment co-camouflaged objects more accurately and completely under different complex scenarios when compared with other comparison methods.
Besides, Fig.~\ref{fig:pr_cosod} provides the PR curves of our proposed method and other CoSOD models, showing the favorable performance of our method.

\begin{table}[tb]
	\centering
	\caption{Ablation study for our proposed modules. The best results are marked in bold.}
	\renewcommand{\arraystretch}{1.5}	\renewcommand{\tabcolsep}{1.85mm}
	\scalebox{0.81}{
		\begin{tabular}{l|cccccc}
			\hline
			\multicolumn{1}{c|}{Model} & \multicolumn{6}{c}{CoCOD8K} \bigstrut\\
			\hline
			B CFE OFS LGR & $S_{\alpha} \uparrow$     & $\mathcal{M} \downarrow$   & $E_{max}\uparrow$ & $F_{max}\uparrow$ & $E_{mean}\uparrow$ & $F_{mean}\uparrow$ \bigstrut\\
			\hline
			$\surd$     & 0.729 & 0.068 & 0.838 & 0.576 & 0.747 & 0.517 \bigstrut[t]\\
			$\surd$ ~$\surd$   & 0.740 & 0.062 & 0.842 & 0.595 & 0.756  & 0.537 \\
			$\surd$ ~$\surd$~ ~~$\surd$ & 0.754 & 0.057 & 0.849 & 0.617  & 0.760 & 0.562 \\
			$\surd$ ~$\surd$~ ~~$\surd$ ~~~ $\surd$ & \textbf{0.789} & \textbf{0.046}  & \textbf{0.855} & \textbf{0.688} & \textbf{0.840} & \textbf{0.670} \bigstrut[b]\\
			\hline
	\end{tabular}}%
	\label{tab:ab_modules}%
\end{table}%

\subsection{Ablation Study} \label{sec:result2}

To verify the effectiveness of our proposed model, the ablation experiments are conducted on the following four aspects: 1) B (dual-branch backbone with element-wise multiplication for feature integration and decoder), 2) B + CFE, 3) B + CFE + OFS, 4) B + CFE + OFS + LGR (Ours). The experimental results are listed in Tab.~\ref{tab:ab_modules}. All ablation experiments use the same parameter settings. Detailed analyses are as follows.

\textbf{Effectiveness of CFE.} 
The collaborative feature extraction module is adapted to capture the common attributes of camouflaged objects, which filter out other non-co-camouflaged objects. From Tab.~\ref{tab:ab_modules}, compared with model B, the addition of the CFE module achieves significant performance improvement. Specifically, the scores of $S_\alpha$, $F_{mean}$, $E_{mean}$ and $\mathcal{M}$ are increased by 1.51\%, 3.87\%, 1.20\% and 8.82\%, respectively.

Inspired by the human observation mechanism, we designed an iterative observation component, \textit{i.e.}, the multi-view feature exploration, in the CFE module. To demonstrate its effectiveness, we test the proposed model with a different number of iterations, shown in Tab.~\ref{tab:ab_cfe_stage}.  
Compared with 0 iterations (\textit{i.e.}, without the multi-view component), adding the multi-view component increases the performance of co-camouflaged object detection and achieves the best performance at two iterations. The performance will deteriorate if continue to increase the number of iterations. We argue the overuse of iterative observation may lead to information interference or misidentification of certain background regions as co-camouflaged objects. Therefore, we set the number of iterations to 2 in our experiments. 

\begin{table}[tb]
	\centering
	\caption{Ablation experiments of iterative observation component (\textit{i.e.}, multi-view feature exploration) in CFE on the CoCOD8K dataset. 0, 1, 2, and 3 denote the number of iterations. The best results are marked in bold.}
	\renewcommand{\arraystretch}{1.25}	\renewcommand{\tabcolsep}{1.35mm}
	\begin{tabular}{c|cccccc}
		\hline
		\multirow{2}[4]{*}{Model} & \multicolumn{6}{c}{CoCOD8K} \bigstrut\\ 
		\cline{2-7} &  $S_{\alpha} \uparrow$     & $\mathcal{M} \downarrow$   & $E_{max}\uparrow$ & $F_{max}\uparrow$ & $E_{mean}\uparrow$ & $F_{mean}\uparrow$ \bigstrut\\
		\hline
		0 & 0.769 & 0.048 & 0.839 & 0.647 & 0.817 & 0.633 \\
		1 & 0.771 & 0.053 & 0.846 & 0.652 & 0.825 & 0.624 \bigstrut[t]\\
		2 & \textbf{0.789} & \textbf{0.046}  & \textbf{0.855} & \textbf{0.688} & \textbf{0.840} & \textbf{0.670} \\
		3 & 0.764 & 0.053 & 0.839 & 0.646 & 0.826 & 0.626 \bigstrut[b]\\
		\hline
	\end{tabular}%
	\label{tab:ab_cfe_stage}%
\end{table}%

\begin{table}[tb]
	\centering
	\caption{Ablation study for the learnable scalar $\gamma$ in OFS. The best results are in bold.}
	\renewcommand{\arraystretch}{1.25}	\renewcommand{\tabcolsep}{1.25mm}
	\begin{tabular}{l|rrrrrr}
		\hline    \multicolumn{1}{c|}{\multirow{2}[4]{*}{Model}} & \multicolumn{6}{c}{CoCOD8K} \bigstrut\\
		\cline{2-7}          & $S_{\alpha} \uparrow$     & $\mathcal{M} \downarrow$   & $E_{max}\uparrow$ & $F_{max}\uparrow$ & $E_{mean}\uparrow$ & $F_{mean}\uparrow$ \bigstrut\\
		\hline
		w/o $\gamma$ & 0.779 & 0.048 & 0.850  & 0.676 & 0.839 & 0.655 \bigstrut[t]\\
		Ours  & \textbf{0.789} & \textbf{0.046} & \textbf{0.855} & \textbf{0.688} & \textbf{0.840}  & \textbf{0.670} \bigstrut[b]\\
		\hline
	\end{tabular}%
	\label{tab:ab_trm}%
\end{table}%

\begin{table}[htbp]
	\centering
	\caption{Ablation experiments of selecting the top $n$ sub-blocks with the largest probability value for local feature exploration in LGR on the CoCOD8K dataset, where $n=\{1,2,3,4\}$. The best results are marked in bold.} 
	\renewcommand{\arraystretch}{1.25}	\renewcommand{\tabcolsep}{1.8mm}
	\begin{tabular}{c|cccccc}
		\hline
		\multirow{2}[4]{*}{$n$} & \multicolumn{6}{c}{CoCOD8K} \bigstrut\\ 
		\cline{2-7} &  $S_{\alpha} \uparrow$     & $\mathcal{M} \downarrow$   & $E_{max}\uparrow$ & $F_{max}\uparrow$ & $E_{mean}\uparrow$ & $F_{mean}\uparrow$ \bigstrut\\
		\hline
		1 & \textbf{0.789} & \textbf{0.046}  & \textbf{0.855} & \textbf{0.688} & \textbf{0.840} & \textbf{0.670} \\
		2 & 0.777 & 0.049 & 0.844 & 0.655 & 0.791 & 0.617 \bigstrut[t]\\
		3 & 0.772 & \textbf{0.046} & 0.842 & 0.649 & 0.788 & 0.615 \bigstrut[b]\\
		4 & 0.776 & 0.049 & 0.842 & 0.654 & 0.789 & 0.614 \bigstrut[b]\\
		\hline
	\end{tabular}%
	\label{tab:ab_lgr}%
\end{table}

\textbf{Effectiveness of OFS.}  
As shown in Tab.~\ref{tab:ab_modules}, compared with B + CFE, the performance of B + CFE + OFS achieves significant performance improvement. 
Specifically, the performance with OFS decreases 8.06\% on $\mathcal{M}$ and increases 1.89\% and 4.66\% on $S_\alpha$ and $F_{mean}$, respectively. 
Besides, we additionally conduct ablation experiments on the learnable scalar $\gamma$, and the results are listed in Tab.~\ref{tab:ab_trm}. In particular, the model with $\gamma$ increases the performance by 1.28\% on $S_\alpha$, 1.37\% on $F_{mean}$ and 4.35\% on $\mathcal{M}$.

\textbf{Effectiveness of LGR.} The local-global refinement module is utilized to refine and enhance the details of co-camouflaged objects. 
As shown in Tab.~\ref{tab:ab_modules}, in contrast to B + CFE + OFS, a further significant performance improvement is achieved with the addition of LGR. In particular, $\mathcal{M}$ decreases 19.30\% (0.011), $E_{mean}$ increases 10.53\% (0.080), and $F_{mean}$ increases 18.15\% (0.102).
In addition, we conduct ablation experiments to verify the effectiveness of selecting the sub-block with the largest probability value as the local clue. 
As shown in Tab.~\ref{tab:ab_lgr}, we test the top $n$ sub-blocks, which are chosen as local feature exploration, which demonstrates that the top 1 sub-block achieves the best performance. 

\begin{figure}[tb]
	\centering
	\includegraphics[width=1\linewidth]{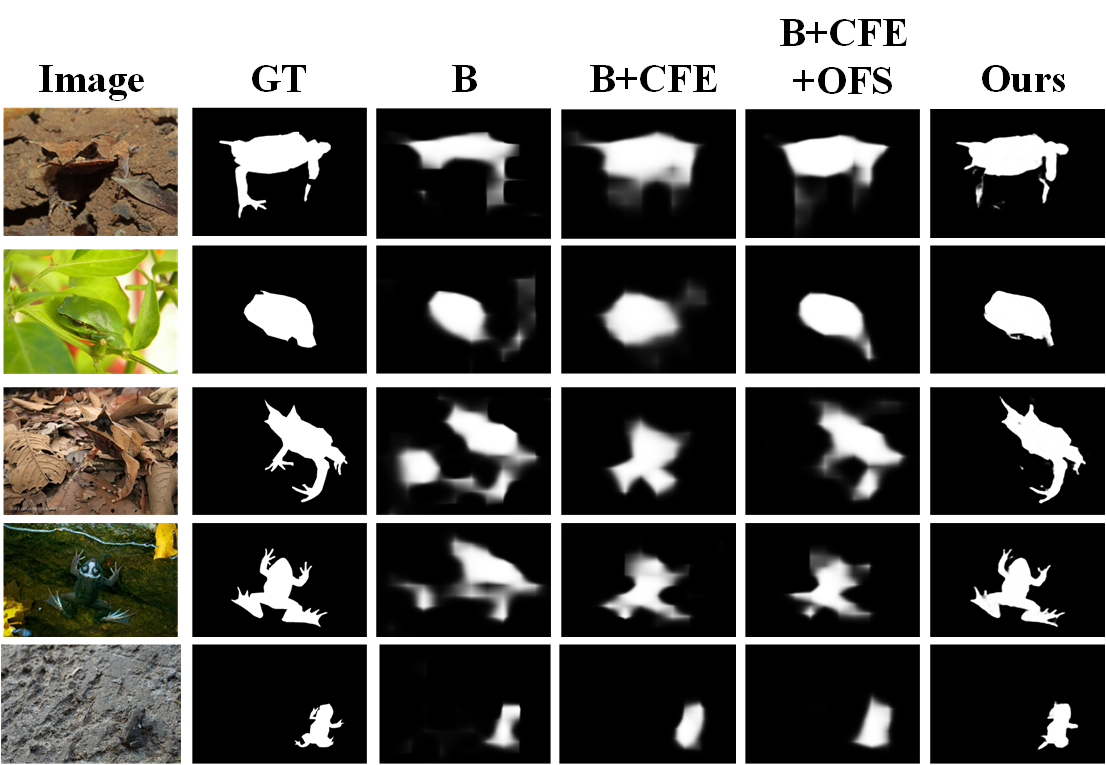}
	\caption{Visual comparison of each proposed module and our final model in ablation experiments.}
	\label{fig:vis_ablation}
\end{figure}

\textbf{Visual comparison of each proposed module.} 
We additionally provide visualization results in Fig.~\ref{fig:vis_ablation} to demonstrate the effectiveness of each proposed module. As shown in the third column, when only leveraging baseline to detect co-camouflaged objects, part of the background is mistaken for objects. Besides, the baseline with the CFE module improves the location of co-camouflaged object regions. Then, we leverage the OFS to capture more camouflaged object cues and integrate dual-branch camouflaged features to improve the object region predictions. Finally, the model with the LGR module complements local details of co-camouflaged objects and provides accurate and complete co-camouflaged object segmentation with fine object outlines. It can be seen that our proposed three modules work together to achieve accurate co-camouflaged object detection.

\textbf{Comparison of different backbones.} 
This section shows the comparison of different backbones on several representative methods on the CoCOD8K dataset, including VGG-16~\cite{karen2015very}, ResNet-50~\cite{he2016deep}, and Res2Net~\cite{gao2019res2net}. In the implementation, we retrain these models based on different backbones using the unified dataset and training parameters. We set the batch size as 10. In addition, we resize the input images to 288$\times$288 for both the training and testing phases. The comparison can be found in Tab.~\ref{tab:backbonecom}.

As can be seen, the proposed model consistently achieves the best overall performance compared to other methods based on VGG-16, ResNet-50, and Res2Net, respectively. For our proposed method, the Res2Net-based version shows superior performance to versions based on the VGG-16 and ResNet-50 backbones.

\begin{table}[tb]
  \centering
  \caption{Comparison of different backbones on some representative methods on the CoCOD8K dataset, including VGG-16~\cite{karen2015very}, ResNet-50~\cite{he2016deep}, and Res2Net~\cite{gao2019res2net}. The best scores are marked in bold.}
  \renewcommand{\arraystretch}{1.25}	\renewcommand{\tabcolsep}{1.25mm}
  \scalebox{0.75}{
    \begin{tabular}{c|c|cccccc}
    \toprule
    Model & Type    
    & $S_{\alpha} \uparrow$     & $\mathcal{M} \downarrow$   & $E_{max}\uparrow$ & $F_{max}\uparrow$ & $E_{mean}\uparrow$ & $F_{mean}\uparrow$ \bigstrut\\
    \midrule
    \multicolumn{8}{c}{VGG-16 as backbone} \\
    \midrule
    \multicolumn{1}{l|}{2022-AAAI-BSANet \cite{zhu2022can}} & COD   & 0.722 & 0.061 & 0.808 & 0.571 & 0.765 & 0.544 \\
    \multicolumn{1}{l|}{2022-IJCAI-BGNet \cite{sun2022boundary}} & COD   & 0.731 & 0.060  & 0.716 & 0.375 & 0.595 & 0.317 \\
    \multicolumn{1}{l|}{2022-CVPR-SegMaR \cite{jia2022segment}} & COD   & 0.624 & 0.133 & 0.734 & 0.432 & 0.587 & 0.339 \\
    \multicolumn{1}{l|}{2021-ICCV-CADC \cite{zhang2021summarize}} & CoSOD & 0.526 & 0.111 & 0.649 & 0.284 & 0.537 & 0.197 \\
    \multicolumn{1}{l|}{2021-CVPR-GCoNet \cite{fan2021group}} & CoSOD & 0.728 & 0.065 & 0.796 & 0.574 & 0.749 & 0.539 \\
    \multicolumn{1}{l|}{2022-CVPR-DCFM \cite{yu2022democracy}} & CoSOD & 0.577 & 0.113 & 0.697 & 0.336 & 0.602 & 0.276 \\
    \multicolumn{1}{l|}{\textbf{Ours}} & -  &\textbf{0.740} & \textbf{0.055} & \textbf{0.814} & \textbf{0.774} & \textbf{0.738} & \textbf{0.574} \\
    
    \midrule
    \multicolumn{8}{c}{ResNet-50 as backbone} \\
    \midrule
    \multicolumn{1}{l|}{2022-AAAI-BSANet \cite{zhu2022can}} & \multicolumn{1}{c|}{COD} & 0.602 & 0.097 & 0.710 & 0.381 & 0.635 & 0.333 \\
    \multicolumn{1}{l|}{2022-IJCAI-BGNet \cite{sun2022boundary}} & \multicolumn{1}{c|}{COD} & 0.629 & 0.089 & 0.720 & 0.417 & 0.653 & 0.375 \\
    \multicolumn{1}{l|}{2022-CVPR-SegMaR \cite{jia2022segment}} & \multicolumn{1}{c|}{COD} & 0.720 & 0.056 & 0.826 & 0.580 & 0.732 & 0.536 \\
    \multicolumn{1}{l|}{2021-ICCV-CADC \cite{zhang2021summarize}} & \multicolumn{1}{c|}{CoSOD} & 0.546 & 0.143 & 0.655 & 0.289 & 0.577 & 0.240 \\
    \multicolumn{1}{l|}{2021-CVPR-GCoNet \cite{fan2021group}} & \multicolumn{1}{c|}{CoSOD} & 0.586 & 0.124 & 0.698 & 0.345 & 0.615 & 0.300 \\
    \multicolumn{1}{l|}{2022-CVPR-DCFM \cite{yu2022democracy}} & \multicolumn{1}{c|}{CoSOD} & 0.741 & 0.058 & 0.818 & 0.595 & 0.747 & 0.552 \\
    \multicolumn{1}{l|}{\textbf{Ours}} & -  & \textbf{0.743} & \textbf{0.052} & \textbf{0.832} & \textbf{0.604} & \textbf{0.766} & \textbf{0.571} \\

    \midrule
    \multicolumn{8}{c}{Res2Net as backbone} \\
    \midrule
    \multicolumn{1}{l|}{2022-AAAI-BSANet \cite{zhu2022can}} & \multicolumn{1}{c|}{COD} &0.755  &0.049  &0.847  &0.638  &0.795  &0.609  \\
    \multicolumn{1}{l|}{2022-IJCAI-BGNet \cite{sun2022boundary}} & \multicolumn{1}{c|}{COD} &0.729  &0.053  &0.828  &0.608  &0.764  &0.577 \\
    \multicolumn{1}{l|}{2022-CVPR-SegMaR \cite{jia2022segment}} & \multicolumn{1}{c|}{COD} &0.587  &0.098  &0.705  &0.375  &0.621  &0.318 \\
    \multicolumn{1}{l|}{2021-ICCV-CADC \cite{zhang2021summarize}} & \multicolumn{1}{c|}{CoSOD} &0.695  &0.065  &0.822  &0.553  &0.761  &0.519 \\
    \multicolumn{1}{l|}{2021-CVPR-GCoNet \cite{fan2021group}} & \multicolumn{1}{c|}{CoSOD} &0.687  &0.066  &0.795  &0.548  &0.739  & 0.522\\
    \multicolumn{1}{l|}{2022-CVPR-DCFM \cite{yu2022democracy}} & \multicolumn{1}{c|}{CoSOD} &0.728  &0.062  &0.815  &0.590  &0.750  &0.548 \\
    \multicolumn{1}{l|}{\textbf{Ours}} & -  &\textbf{0.789} & \textbf{0.046}  & \textbf{0.855} & \textbf{0.688} & \textbf{0.840} & \textbf{0.670} \\

    \bottomrule
    \end{tabular}
    }
  \label{tab:backbonecom}
\end{table}

\textbf{Effectiveness of loss function.} Tab.~\ref{tab:losscom} shows the performance of different loss functions on the CoCOD8K dataset, including the BCE-only loss function and the loss function we used (\textit{i.e.}, BCE, and IoU). 
Collaborative camouflaged object detection aims to jointly segment the same camouflaged object (or camouflaged objects of the same class) in multiple images. Actually, most of the inputs contain abundant background noises, whose texture, color or shape is similar to collaborative camouflaged objects. 
When the number of background elements is much larger than the number of foreground elements, the loss function component in the background element will dominate, and it is easy to make the model biased toward the background. Benefiting from the strong convergence capability of the IoU function in the small object detection task, we adopt the IoU loss and the cross-entropy loss to jointly constrain the model. We can see that the joint loss function achieves better performance than the BCE-only loss function. The introduction of the IoU loss function can facilitate the convergence of the model and slightly improve the detection performance.

\begin{table}[tb]
\centering
\caption{Ablation study of the loss function on the CoCOD8K dataset. The best scores are marked in bold.} 
\renewcommand{\arraystretch}{1.2}	
\renewcommand{\tabcolsep}{1.55mm}
    \begin{tabular}{c|cccccc}
    \toprule
    \multirow{1}[4]{*}{Loss}  &  $S_{\alpha} \uparrow$     & $\mathcal{M} \downarrow$   & $E_{max}\uparrow$ & $F_{max}\uparrow$ & $E_{mean}\uparrow$ & $F_{mean}\uparrow$ \\
    \midrule
    BCE & \textbf{0.789} & 0.048  & 0.850 & 0.683 & 0.831 & 0.664 \\ \hline
    Ours & \textbf{0.789} & \textbf{0.046}  & \textbf{0.855} & \textbf{0.688} & \textbf{0.840} & \textbf{0.670}  \bigstrut[t]\\
    \bottomrule
    \end{tabular}%
    \label{tab:losscom}%
\end{table}

\textbf{Model complexity.} This section provides the comparison of the proposed BBNet and other methods in terms of the number of parameters, model size (FLOPs), and FPS. The details are listed in Tab.~\ref{tab:flops}. We can see that our proposed model is average-level on three metrics compared to other models. We can see that our method achieves a good balance in model complexity, parameter amount, and inference speed. However, to be frank, there is still room for model improvement, such as lightweight models, to enhance the model efficiency and reduce the model complexity and parameters, which is the focus of our future work.

\begin{table}[tb]
  \centering
  \caption{Comparison of the number of parameters (Params), FLOPs and FPS. The evaluation complies with the settings in the respective original papers. The best results are marked in bold.}
  \renewcommand{\arraystretch}{1.15}	\renewcommand{\tabcolsep}{2.15mm}
    \scalebox{0.91}{\begin{tabular}{cc|ccc}
    \toprule
    \multicolumn{1}{c|}{Model} & \multicolumn{1}{c|}{Type} & FLOPs & Params & FPS \bigstrut\\
    \midrule
    \multicolumn{1}{l|}{2020-CVPR-SINet~\cite{fan2020camouflaged}} & \multicolumn{1}{c|}{COD} & 333.065G & 29.737M & 29.7  \bigstrut[t]\\
    \multicolumn{1}{l|}{2021-CVPR-JCOOD~\cite{li2021uncertainty}} & \multicolumn{1}{c|}{COD} & 1.752T & 74.988M & 16.0  \\
    \multicolumn{1}{l|}{2021-CVPR-RankNet~\cite{lv2021simultaneously}} & \multicolumn{1}{c|}{COD} & 195.161G & 55.450M & \textbf{55.7} \\
    \multicolumn{1}{l|}{2021-IJCAI-C2FNet \cite{sun2021context}} & \multicolumn{1}{c|}{COD} & 392.490G & 15.829M & 23.4  \\
    \multicolumn{1}{l|}{2021-CVPR-PFNet\cite{mei2021camouflaged}} & \multicolumn{1}{c|}{COD} & 333.168G & 29.078M & 27.0  \\
    \multicolumn{1}{l|}{2022-TPAMI-SINet-v2~\cite{fan2022concealed}} & \multicolumn{1}{c|}{COD} & 220.273G & \textbf{15.603M} & 41.5  \\
    \multicolumn{1}{l|}{2022-CVPR-SegMaR \cite{jia2022segment}} & \multicolumn{1}{c|}{COD} & 836.995G & 64.731M & 37.8  \\
    \multicolumn{1}{l|}{2022-PR-ERRNet \cite{ji2022fast}} & \multicolumn{1}{c|}{COD} & 523.540G & 32.053M & 51.9  \\
    \multicolumn{1}{l|}{2022-WACV-OCENet \cite{liu2022modeling}} & \multicolumn{1}{c|}{COD} & 660.245G & 38.139M & 33.1  \\
    \multicolumn{1}{l|}{2022-IJCAI-BGNet \cite{sun2022boundary}} & \multicolumn{1}{c|}{COD} & 533.487G & 23.198M & 31.5  \\
    \multicolumn{1}{l|}{2022-KBS-BgNet \cite{chen2022boundary}} & \multicolumn{1}{c|}{COD} & 519.547G & 28.289M & 33.2  \\
    \multicolumn{1}{l|}{2022-AAAI-BSANet \cite{zhu2022can}} & \multicolumn{1}{c|}{COD} & 132.568G & 28.792M & 26.0  \\
    \multicolumn{1}{l|}{2020-ECCV-GICD \cite{zhang2020gradient}} & \multicolumn{1}{c|}{CoSOD} & 329.563G & 25.451M & 49.1  \\
    \multicolumn{1}{l|}{2020-NeurIPS-ICNet \cite{jin2020icnet}} & \multicolumn{1}{c|}{CoSOD} & 240.238G & 25.395M & 44.5  \\
    \multicolumn{1}{l|}{2021-TPAMI-CoEGNet \cite{fan2021re}} & \multicolumn{1}{c|}{CoSOD} & 1.964T & 45.558M & 33.2  \\
    \multicolumn{1}{l|}{2021-ICCV-CADC \cite{zhang2021summarize}} & \multicolumn{1}{c|}{CoSOD} & \textbf{20.194G} & 379.372M & 11.4  \\
    \multicolumn{1}{l|}{2021-CVPR-GCONet \cite{fan2021group}} & \multicolumn{1}{c|}{CoSOD} & 209.351G & 25.762M & 27.8  \\
    \multicolumn{1}{l|}{2022-CVPR-DCFM \cite{yu2022democracy}} & \multicolumn{1}{c|}{CoSOD} & 436.961G & 67.415M & 32.3  \bigstrut[b]\\
    \hline
    \multicolumn{2}{l|}{Ours} & 294.952G & 24.305M & 41.7  \bigstrut\\
    \bottomrule
    \end{tabular}
    }%
  \label{tab:flops}%
\end{table}%

\section{Conclusion}
\label{sec:conclusions}

In this paper, we propose the first work for collaborative camouflaged object detection. To achieve this goal, we construct a large-scale challenging CoCOD8K dataset, which contains 8,528 high-resolution camouflaged images covering 5 super-classes and 70 sub-classes. 
Then, we design a simple but effective bilateral-branch network (BBN), which effectively integrates intra-image and inter-image camouflaged cues for accurate co-camouflaged object detection from a group of relevant images. 
Besides, we provide a comprehensive benchmark on 18 existing cutting-edge COD and CoSOD methods under 5 widely used evaluation metrics. The quantitative and qualitative experiments demonstrate the effectiveness of the proposed BBNet and its superior performance to other competitors. 
We hope the studies presented in this work (\textit{i.e.}, dataset and baseline) will offer the community an opportunity to explore more and spark novel ideas in this field. 

\section{Acknowledgments}
The work is supported by the Natural Science Foundation of Heilongjiang Province under No.~LH2022F005 (Research on Visual Camouflaged Object Detection for Multi-Modal and Multi-Target Data).

\bibliographystyle{IEEEtran}
\bibliography{refs}

\end{document}